\documentclass{article}

\usepackage[preprint, nonatbib]{neurips_2020}

\usepackage[utf8]{inputenc} 
\usepackage[T1]{fontenc}    
\usepackage{hyperref}       
\usepackage{url}            
\usepackage{booktabs}       
\usepackage{amsfonts}       
\usepackage{nicefrac}       
\usepackage{microtype}      

\usepackage[square,numbers]{natbib}

\usepackage{amsmath,amsfonts,bm, bbm}
\usepackage{amssymb}
\usepackage{mathtools}

\def\eqref#1{equation~\ref{#1}}

\def\1{\bm{1}}

\def\rvx{{\mathbf{x}}}
\def\rvy{{\mathbf{y}}}
\def\rvz{{\mathbf{z}}}

\def\vz{{\bm{z}}}

\def\mI{{\bm{I}}}

\DeclareMathAlphabet{\mathsfit}{\encodingdefault}{\sfdefault}{m}{sl}
\SetMathAlphabet{\mathsfit}{bold}{\encodingdefault}{\sfdefault}{bx}{n}

\usepackage{tikz}
\usetikzlibrary{shapes.geometric}

\usepackage{caption}
\usepackage{subcaption}
\usepackage{xcolor}

\usepackage[ruled,vlined]{algorithm2e}
\newenvironment{algo}[1][htb]
  {
  \begin{algorithm}[#1]%
  }{\end{algorithm}}
  
\usepackage{chngcntr}

\newcommand{\jw}[1]{\textcolor{purple}{\bf [jack: #1]}}

\newcommand{\wn}[1]{\textcolor{blue}{\bf [wn: #1]}}

\title{An Improved Semi-Supervised VAE for Learning Disentangled Representations}

\author{%
  Weili Nie$^{1}$\thanks{Equal contribution.}\hspace{25pt} 
  Zichao Wang$^{1\hspace{0.6pt}*}$\hspace{25pt} 
  Ankit B. Patel$^{1,2}$\hspace{25pt} 
  Richard G. Baraniuk$^{1}$ \\
  $^{1}$Department of Electrical and Computer Engineering, Rice University\\
  $^{2}$ Baylor College of Medicine\\
  \texttt{\{wn8, zw16, abp4, richb\}@rice.edu} \\
}

\begin{document}

\maketitle

\begin{abstract}
Learning interpretable and disentangled representations is a crucial yet challenging task in representation learning. 
In this work, we focus on semi-supervised disentanglement learning and extend~\cite{locatello2019-semi-disentangling} by introducing another source of supervision that we denote as label replacement.
Specifically, during training, we replace the inferred representation associated with a data point with its ground-truth representation whenever it is available. 
Our extension to~\cite{locatello2019-semi-disentangling} is 
theoretically inspired by our proposed general framework of semi-supervised disentanglement learning in the context of VAEs which naturally motivates the supervised terms commonly used in existing semi-supervised VAEs (but not for disentanglement learning), e.g.,~\cite{kingma_semi}.
Extensive experiments on synthetic and real datasets demonstrate both quantitatively and qualitatively the ability of our extension to significantly and consistently improve disentanglement with very limited supervision.

\end{abstract}
\section{Introduction}

\label{introduction}

Learning a disentangled representation has recently emerged as a foundational task in machine learning. For a given data point, its representation (or {\it label}, in the form of a multi-dimensional vector) is ``disentangled'' when each dimension of the label independently controls the variation of one single attribute (factor of variation) of the data point~\cite{locatello2019challenging, szabo2017challenges}.
Two tasks are of typical interest in disentanglement learning. The {\it encoding} tasks entails {\it inferring} the label that represents the true factors of variation given a data point. These inferred labels can serve as interpretable and efficient summaries of data points, which can be useful for many downstream tasks~\cite{bengio2013representation}. The {\it decoding} task entails {\it generating}, given a label as input, a data point whose attributes corresponds exactly to what the input label specifies. Such decoding allows the generation of data points with the exact factors of variation in a controlled and interpretable manner, which has a wide range of real-world applications including speech synthesis~\cite{habib2019semi}, fairness~\cite{creager2019flexibly}, and computer graphics~\cite{aumentado2019geometric}.

Variational auto-encoders (VAEs) have attracted increasing attention for disentanglement learning, because of their capability to jointly learn models for both the encoding and decoding tasks and because of the feasibility to impose structural constraints on them to encourage disentanglement.
Prior work has largely focused on {\it unsupervised} disentanglement learning,
in which the ground-truth label associated with each data point is unavailable to the model. However, in the unsupervised setting, a model is non-identifiable: there can exist multiple models capable of producing distinct but equally valid code of a data point~\cite{locatello2019challenging, khemakhem2019variational}. This goes against the goal of disentanglement learning because if multiple different labels exist for the same data point, the semantic meanings of each dimension of the different labels are not consistent and are thus no longer interpretable. 

The above observations suggest that some form of {\em supervision} using the ground-truth labels has the potential to improve disentanglement learning. However, collecting ground-truth labels for all data points is costly and labor-intensive
This naturally leads to the \textit{semi-supervised} setting, where we assume the ground-truth labels are known for a {\it very limited} number of data points.
Unfortunately, to date the investigation of disentanglement learning in the semi-supevised setting remains scarce.
Among the few existing works,~\citet{locatello2019-semi-disentangling} have shown that simply adding a ``label loss'', which minimizes the difference between the inferred labels and the limited available ground-truth labels, to unsupervised VAEs leads to improved disentanglement compared to their unsupervised counterparts.
However, it is not clear to what extent can (limited) ground-truth labels effectively improve disentanglement learning.

{\bf Contributions.}~ 
In this paper, we study semi-supervised disentanglement learning in the context of VAEs. Specifically, we extend~\cite{locatello2019-semi-disentangling} by more effectively exploiting information in the limited labeled data. This is achieved by supplying the ground-truth labels to the decoder, whenever they are available, in order to regularize the data generation process. 
Our extension to~\cite{locatello2019-semi-disentangling} is theoretically inspired by our general formulation for semi-supervised disentanglement learning that unifies both label replacement and the label loss commonly employed in existing semi-supervised VAEs.
Extensive experiments on multiple datasets demonstrate the superior performance of our label replacement extension to baseline models in~\cite{locatello2019-semi-disentangling} without label replacement.




\section{Preliminaries}

Consider a generative model with a multivariate latent variable $\bm{\xi}$, usually sampled from a simple factorized prior distribution $p(\bm{\xi})$,
and an observation sampled from the conditional distribution $p(\rvx|\bm{\xi})$, where $d > 1$.
The goal of disentangled representation learning is to learn a presentation $r(\rvx)$ that separates different factors of variation in the observation $\rvx$. Thus, a change in each dimension of the learned representation $r(\rvx)$ is only caused by the change in a dimension of $\bm{\xi}$.

{\bf    Unsupervised Disentanglement Learning with VAEs.}~
Many state-of-the-art unsupervised disentanglement methods are VAE-based models \citep{locatello2019challenging}. 
VAEs typically assume that the prior $p(\bm{\xi})$ is a simple distribution, such as an isotropic Gaussian. 
The conditional distribution $p_\theta(\rvx | \bm{\xi})$ is usually parametrized by a deep neural network called the \textit{decoder}. Similarly, the posterior $p(\bm{\xi} | \rvx)$ is approximated with a variational distribution $q_\phi(\bm{\xi} | \rvx)$, which is also parametrized by a deep neural network called the \textit{encoder}. Here, we denote by $\theta$ and $\phi$ the parameters of the encoder and decoder, respectively.
Most unsupervised disentanglement methods with VAEs regularize the average evidence lower-bound (ELBO) by minimizing the total correlation \citep{betatcvae}, and thus the unsupervised loss can be summarized as
\begin{align} \label{unsup}
    \mathcal{L}_{\rm unsup} = \mathbb{E}_{\rvx} [-{\rm ELBO}] 
    + \gamma_{\rm tc} \mathbb{E}_{\rvx}[R_u(q_{\phi}(\bm{\xi} | \rvx))]
\end{align}
where 
\begin{align} \label{elbo}
     {\rm ELBO} \triangleq \mathbb{E}_{q_{\phi}(\bm{\xi}| \rvx)} [\log p_{\theta}(\rvx | \bm{\xi}) ] - D_{\rm KL} (q_{\phi}(\bm{\xi}| \rvx) || p(\bm{\xi}))
\end{align}
and $\gamma_{\rm tc}$ is the weight of the total correlation term, and the choice of the function $R_u(\cdot): \mathbb{R} \to \mathbb{R}$ depends on the specific methods \citep{betavae, factorvae, betatcvae}.

{\bf    Semi-Supervised Disentanglement Learning with VAEs.}~
Prior work on semi-supervised disentanglement learning considers a dataset $\mathcal{D}$ consisting of a large set of unlabeled data $\mathcal{P}_U$ and a small set of labeled data $\mathcal{P}_L$, where $\mathcal{D} = \mathcal{P}_L \cup \mathcal{P}_U$ and $|\mathcal{P}_L| \ll |\mathcal{P}_U|$. 
Because now (limited) ground-truth labels are available, the graphical model in the semi-supervised setting becomes different from its unsupervised counterpart. 
Typically~\cite{locatello2019-semi-disentangling, nie2020semi}, the latent variable $\bm{\xi}$ is partitioned into two portions $\bm{\xi} = (\rvy, \rvz)$, where the {\it label} $\rvy$ represents the considered ground-truth factors of variation associated with a data point and the {\it nuisance} $\rvz$ represents other factors of variation that $\rvy$ does not capture.
This implies that to ensure disentangled representations, $\rvy$ and $\rvz$ should be assumed to be \textit{conditionally independent}. Thus,
the variational posterior parametrized by the encoder is factorized as
\begin{align} \label{factor_q}
    q_{\phi}(\bm{\xi} | \rvx) = q_{\phi}(\rvy | \rvx) q_{\phi}(\rvz | \rvx).
\end{align}

To use supervision for better disentanglement, prior work typically incorporates a so-called \textit{label loss} as a supervised regularization term into $\mathcal{L}_{\rm unsup}$ (Eq.~\ref{unsup}) \cite{siddharth2017learning, locatello2019-semi-disentangling}. Therefore, the baseline semi-supervised loss is given by 
\begin{align} \label{semi-sup}
    \mathcal{L}_{\rm baseline} = \mathcal{L}_{\rm unsup} + \gamma_{\rm lb} \mathbb{E}_{\rvx, \rvy \sim \mathcal{P}_L }[R_s(q_{\phi}(\bm{\rvy} | \rvx)),
\end{align}
where $\gamma_{\rm lb}$ denotes the weight of label loss, and the function $R_s(\cdot)$ is decided by the type of label loss, such as the binary cross-entropy loss or the mean square error (MSE).  

From the the above loss in Eq.~\ref{semi-sup}, we can see that the baseline semi-supervised method imposes the supervision to only guide the {\it encoder} for reconstructing the labels, in an intuitive yet relatively ad-hoc way. The lack of a principled framework for semi-supervised disentangled VAEs may make it suboptimal in using labeled data for disentanglement learning.

\begin{figure}
    \centering
    \captionsetup[subfigure]{justification=centering}
    \begin{subfigure}[t]{0.45\linewidth}
    \centering
    \includegraphics[width=0.5\linewidth]{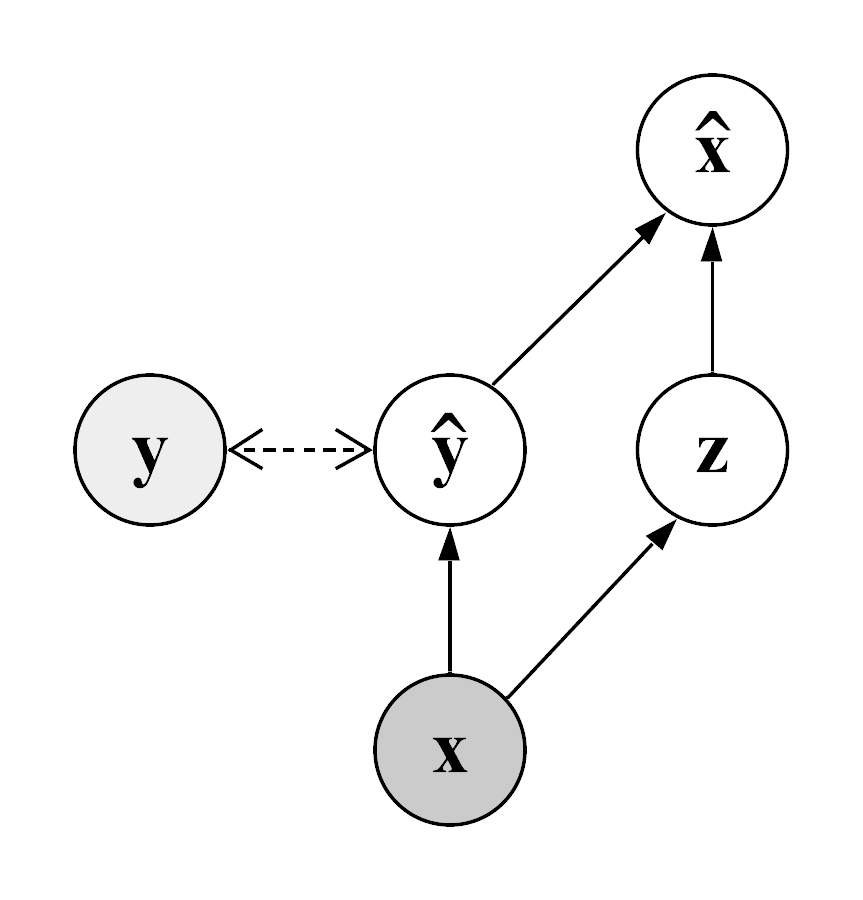}
    \caption{Semi-supervised disentanglement VAE~\cite{locatello2019-semi-disentangling}}
    \label{fig:semi-dis-vae}
    \end{subfigure}
    \begin{subfigure}[t]{0.45\linewidth}
    \centering
    \includegraphics[width=0.5\linewidth]{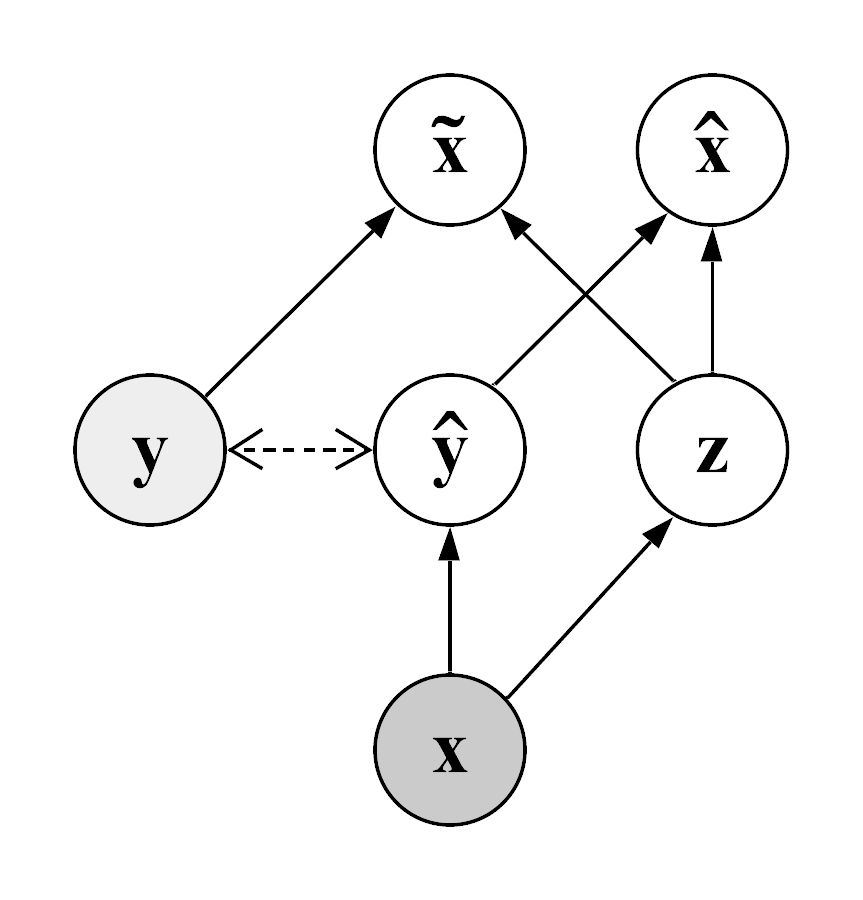}
    \caption{LaRVAE}
    \label{fig:LaRVAE}
    \end{subfigure}
    \caption{\small Illustration of the semi-supervised disentanglement VAEs in~\cite{locatello2019-semi-disentangling} (Fig.~\ref{fig:semi-dis-vae}) and our label replacement extension (Fig.~\ref{fig:LaRVAE}). 
    In Fig.~\ref{fig:LaRVAE}, we use $\widetilde{\rvx}$ and $\widehat{\rvx}$ to differentiate the two data points generated from the {\it ground-truth} label $\rvy$ and from the {\it inferred} label $\widehat{\rvy}$. 
    Light grey indicates only limited ground-truth labels $\rvy$'s are available while dark grey indicates all data points $\rvx$'s are fully available and observed. The dotted, double-sided arrow represents the label reconstruction loss $\mathcal{L}_{\rm recon}$ (see Eq.~\ref{eq:semi-general}).
    }
    \label{fig:models}
\end{figure}

\section{Method}
We now derive our proposed label replacement extension to the semi-supervised disentanglement VAEs in~\cite{locatello2019-semi-disentangling}  from a new, general formulation for semi-supervised disentanglement learning. 
Our general formulation naturally incorporates both the introduced label replacement term and the label loss in a principled manner.

\subsection{The label Replacement Disentanglement VAE Objective Function}
\label{sec:semi-sup}
%

We consider a generalized negative log-likelihood (NLL) objective for semi-supervised disentanglement learning in the context of VAEs. 
To do so, we directly incorporate a joint NLL term into the unsupervised objective $\mathcal{L}_{\rm unsup}$ in Eq.~\ref{unsup}, resulting in
\begin{align} \label{nll_orig}
        \mathcal{L}_{\rm semi} = & \mathcal{L}_{\rm unsup}
        + \gamma\mathbb{E}_{\rvx, \rvy\sim\mathcal{P}_{L}}[-{\rm log}\,p_\theta(\rvx, \rvy)]\,,
\end{align}
where the joint NLL conveys all the supervised information provided by the labeled data $(\rvx, \rvy)$. The hyperparameter $\gamma$ controls the weight of the supervised term. Note that Eq.~\ref{nll_orig} is more general than Eq.~\ref{semi-sup} because the supervised regularization term in Eq.~\ref{nll_orig} considers the joint distribution rather than a conditional distribution and does not invoke variational approximation. This supervised joint NLL term will be key to the subsequent derivation and discussion of LaRVAE.

We then decompose the log of the joint distribution ${\rm log}\,p_\theta(\rvx, \rvy)$ into
\begin{align} 
    {\rm log}\,p_\theta(\rvx, \rvy) &= \lambda {\rm log}\,p_\theta(\rvy|\rvx) p_\theta(\rvx) + (1 - \lambda){\rm log}\, p_\theta(\rvx|\rvy) p_\theta(\rvy)\label{bayes-raw} \,.
\end{align}
This decomposition naturally combines loss terms for two tasks: 1) the decoding task $p_\theta(\rvx|\rvy)$ given the label $\rvy$, and 2) the encoding task $p_\theta(\rvy|\rvx)$ given the data point $\rvx$. 
Thus, via the hyperparameter $\lambda$, the joint distribution balances the interplay between the above two tasks for better disentanglement.

%
Finally, we obtain the objective of our label replacement extension of semi-supervised disentanglement VAEs by substituting Eq.~\ref{bayes-raw} into Eq.~\ref{nll_orig} and ignoring a trivial scaling factor:
\begin{align}
    \mathcal{L}_{\rm semi} \approx & \mathcal{L}_{\rm unsup}
    + \alpha\underbrace{\mathbb{E}_{\rvx, \rvy\sim\mathcal{P}_L}[-{\rm log}\, {q}_\phi(\rvy|\rvx)]}_{\mathcal{L}_{\rm recon}}
    + \tau\underbrace{\mathbb{E}_{\rvx,\rvy\sim\mathcal{P}_L}[-{\rm log}\,p_\theta(\rvx|\rvy)]}_{\mathcal{L}_{\rm rep}}\,,
\label{eq:semi-general}
\end{align}
where $\alpha = \frac{\lambda \gamma}{1 + \lambda \gamma}$ and $\tau = \frac{ (1-\lambda) \gamma}{1 + \lambda \gamma}$.
Note that we have approximated $p_\theta(\rvy|\rvx)$, which is intractable to compute, with the variational distribution ${q}_\phi(\rvy|\rvx)$.
We have also omitted the $p_\theta(\rvy)$ term, because it is usually assumed to be a standard Gaussian and thus does not involve any model parameters. 
A detailed derivation is available in the Supplementary Material.

By starting with a more general supervised loss term (Eq.~\ref{nll_orig}), 
we have now introduced two supervised regularization terms, $\mathcal{L}_{\rm recon}$ and $\mathcal{L}_{\rm rep}$, in a principled manner. 
$\mathcal{L}_{\rm recon}$ is the label reconstruction loss commonly added in existing semi-supervised VAEs but in an ad-hoc way. 
$\mathcal{L}_{\rm rep}$ is a novel and effective supervised loss that we call {\it label replacement loss}, which we describe next.

\subsection{The Label Replacement Loss $\mathcal{L}_{\rm rep}$}
Intuitively, whenever a labeled data point $\rvx, \rvy$ from $\mathcal{P}_{L}$ is available, we {\it replace} the inferred label $\hat{\rvy}$ with the ground-truth label $\rvy$ as the input to the decoder to generate $\tilde{\rvx}$.
We then use $\tilde{\rvx}$ to regularize the image reconstruction process in the decoder.
See pathway leading to $\tilde{\rvx}$ in Fig.~\ref{fig:LaRVAE} for an illustration.
This simple way of exploiting the labeled data is absent in existing semi-supervised disentanglement VAEs which focus on only regularizing the encoder instead.
We discuss how LaRVAE relates, differs, and extends prior work in more detail in Section 4.

At first glance, since $\mathcal{L}_{\rm rep}$ only concerns the {\it decoder}, one may wonder why $\mathcal{L}_{\rm rep}$ would improve the ability of learning disentangled representations, a main feature in the encoder.
This may explain why previous work consider only regularizing the encoder with the label loss in Eq.~\ref{semi-sup}. 
However, we note that the encoder and decoder are trained jointly in VAE, and that the encoder and decoder are connected via the inferred label $\hat{\rvy}$ when ground-truth label is not available. Thus, our hypothesis is that the better decoder would still have a significant effect on the encoder for a improved disentanglement performance. We will demonstrate the positive impact of adding $\mathcal{L}_{\rm rep}$ with extensive empirical evidence in Section 5.

\subsection{Implementation Details}

{\bf    Computing $\mathcal{L}_{\rm recon}$ and $\mathcal{L}_{\rm rep}$.} 
Given the factorized form of $q_{\phi}(\bm{\xi} | \rvx)$ in Eq. \ref{factor_q}, we can approximate the posteriors ${q}_{\phi}(\rvy | \rvx)$ and ${q}_{\phi}(\rvz | \rvx)$ separately. First, we use a conditional Gaussian to parametrize the approximate posterior ${q}_{\phi}(\rvy | \rvx)$ of $\mathcal{L}_{\rm recon}$ in Eq. \ref{eq:semi-general} as 
${q}_{\phi} (\rvy | \rvx) = \mathcal{N} ({\bm{\mu}}_{y_{\phi}}(\rvx), \sigma^2 \mI )$, 
where $\bm{\mu}_{y_{\phi}}(\rvx)$ 
is one output of the encoder, and the the variance is set to a constant $\sigma^2$ for simplicity.
Therefore, the label reconstruction loss $\mathcal{L}_{\rm recon}$ is computed as 
\begin{align}
     \mathcal{L}_{\rm recon} = \alpha' \mathbb{E}_{\rvx,\rvy\sim\mathcal{P}_L}[\| \bm{\mu}_{y_{\phi}}(\rvx) - \rvy \|^2] \,, \label{eq:recon}
\end{align}
where $\alpha'> 0$ absorbs other constant terms independent of the model parameters $\theta$ and $\phi$.
The label replacement loss $\mathcal{L}_{\rm rep}$ is computed via 
\begin{align} 
        \mathcal{L}_{\rm rep} 
    &= -\mathbb{E}_{\rvx,\rvy\sim\mathcal{P}_L, \rvz\sim q_\phi(\rvz|\rvx)} \left[{\rm  log}\,p_\theta(\rvx|\rvy,\rvz)\right] 
     - \mathbb{E}_{\rvx\sim\mathcal{P}_L} [ D_{\rm  KL}(q_\phi(\rvz|\rvx) || p(\rvz))]. 
     \label{eq:rep} 
\end{align} 
In the above equation, we use another conditional Gaussian to parametrize the conditional data likelihood $p_{\theta} (\rvx | \rvy,\rvz) = \mathcal{N} ({\bm {\mu}}_{\theta}(\bm{\xi}), \sigma^2 \mI )$, where
$\bm{\mu}_\theta({\bm{\xi}})$ is the output of the decoder with the concatenation of $\rvy$ and $\rvz$ as input. 
Also, similar to ${q}_{\phi}(\rvy | \rvx)$, we parametrize the posterior ${q}_{\phi}(\rvz | \rvx)$ in Eq. \ref{eq:rep} as ${q}_{\phi} (\rvy | \rvx) = \mathcal{N} ({\bm{\mu}}_{z_{\phi}}(\rvx), \sigma^2 \mI )$, where $\bm{\mu}_{z_{\phi}}(\rvx)$ 
is another output of the encoder.
The detailed derivations of Eqs.~\ref{eq:recon} and~\ref{eq:rep} are available in the Supplementary Material.
The remaining terms in Eq.~\ref{eq:semi-general} are straightforward and are the same as in~\cite{locatello2019-semi-disentangling}.

{\bf    Training.}
In each iteration during training, we sample a batch $\{\rvx\}_i^{K}$ from $\mathcal{D}$ and another batch of $\{\rvx_i,\rvy_i\}$ from $\mathcal{P}_L$. This is to ensure that the model receives sufficient labeled data. Note that data points in the first batch can also appear in the second batch because $\mathcal{P}_L\subset\mathcal{D}$. We then compute the loss Eq.~\ref{eq:semi-general} using Eqs.~\ref{eq:recon} and~\ref{eq:rep} and update model parameters $\theta$ and $\phi$ using first-order optimization techniques (we use Adam~\cite{adam} in experiments). 
Pleaase see Table 1 in Supplementary Material for a summary of the LaRVAE training procedure.



\section{Related Work}

{\bf Relation to (Disentangled) Semi-Supervised VAEs.}~
Our work builds on and extends~\cite{locatello2019-semi-disentangling}. Specifically, we adopt the graphical model of~\cite{locatello2019-semi-disentangling} which incorporates the label $\rvy$ directly as part of the latent variable so that $\rvz$ does not depend on $\rvy$; see Fig.~\ref{fig:semi-dis-vae} for an illustration. 
This is in contrast to a number of existing works on semi-supervised learning using VAEs~\cite{kingma_semi,siddharth2017learning,de2018semi} where the latent variable is conditioned on $\rvy$. 
Our model structure and that in~\cite{locatello2019-semi-disentangling} is appropriate for studying disentanglement for two reasons. 
First, our graphical model setup incorporates interpretability directly into the latent variables, because part of the latent variable contains $\rvy$ which represents the factors of variations and is easily interpretable. 
In contrast, the latent variable in the graphical model in~\cite{kingma_semi} is not interpretable because no structural restrictions are imposed.
Second, because $\rvy$ is part of the latent variable, our graphical model is compatible with a number of unsupervised disentanglement VAEs~\cite{betavae,betatcvae,factorvae} that apply disentanglement regularizations to the latent variable. Thus, we can leverage these VAE models in our extension to improve disentanglement. In contrast, the graphical model in~\cite{kingma_semi} is not compatible with existing unsupervised disentanglement VAEs.

LaRVAE also easily extends to semi-supervised learning in a more general setting.
This is because the unsupervised loss term $\mathcal{L}_{\rm unsup}$ in LaRVAE objective is a lower bound of ELBO (see Eq.~\ref{unsup}) and thus a lower bound of the marginal NLL term $\mathbb{E}_{\rvx}[p_\theta(\rvx)]$ in the general semi-supervised learning problem formulation. Therefore, we can replace $\mathcal{L}_{\rm unsup}$ with ${\rm ELBO}$ in the LaRVAE objective which generalizes LaRVAE to the generic semi-supervised learning setting. Because the present paper focuses on disentanglement learning, we defer the investigation of applying LaRVAE to general semi-supervised learning to future work.

{\bf    Other Related Work on Disentanglement Learning.}~ 
The majority of disentanglement learning literature takes an unsupervised learning approach under the VAE framework, notably including $\beta$-VAE~\cite{betavae}, FactorVAE~\cite{factorvae} and $\beta$-TC-VAE~\cite{betatcvae}. These works regularize the inferred factors by decomposing the KL divergence term in different ways (also see similar decompositions in~\cite{infovae, structured-vae, dd}). These works also propose novel evaluation metrics that we continue to use in our work. Some other works impose regularizations in different ways. For example,~\cite{csvae, mutual-vae} propose explicit mutual information-like regularization term that to encourage the information about the input data points to remain on the desired (subset of) factors. 
Other models such as GANs~\cite{infogan, infogan-cr} instead of VAEs were also considered in prior work.
Although these approaches show promise,~\cite{locatello2019challenging, khemakhem2019variational} demonstrates that unsupervised disentanglement models are not identifiable. These results suggest that supervision is necessary and motivate us to consider semi-supervised setting using limited ground-truth labels.


Another line of research, although limited, uses explicit supervision for disentanglement learning. Some combines both VAE and GAN framework for semi-supervised disentangled representation learning~\cite{zhang2019adversarial}, with application to human pose estimation~\cite{de2018semi}. ~\cite{Mathieu16a} adds an adversarial loss on the labels, although the setting is fully supervised. 
Our work contributes to this line of research by developing an effective way to exploit label information to improve disentanglement learning using VAEs in the semi-supervised setting.

\section{Experiments}


We perform extensive experiments to demonstrate the effectiveness of our label replacement extension to~\cite{locatello2019-semi-disentangling} for semi-supervised disentanglement learning. From now on, we refer to our extension as LaRVAE which stands for label replacement VAE. We first quantitatively show that, on various (limited) numbers of available ground-truth labels, LaRVAE outperforms various baseline semi-supervised disentanglement VAEs. 
We also investigate the sensitivity of LaRVAE to different hyperparameter. 
Finally, we qualitatively show that LaRVAE generates samples of higher quality as compared to the baseline, by using the label traversals. 
More details on the datasets, data preprocessing procedures, model architectures and experiment setups are available in the Supplementary Material.

\subsection{Quantitative Evaluations}

\begin{figure} [t]
  \centering
  \begin{subfigure}[b]{1\textwidth}
		\centering
		\small
		\makebox[\textwidth][c]{\includegraphics[width=1.025\linewidth]{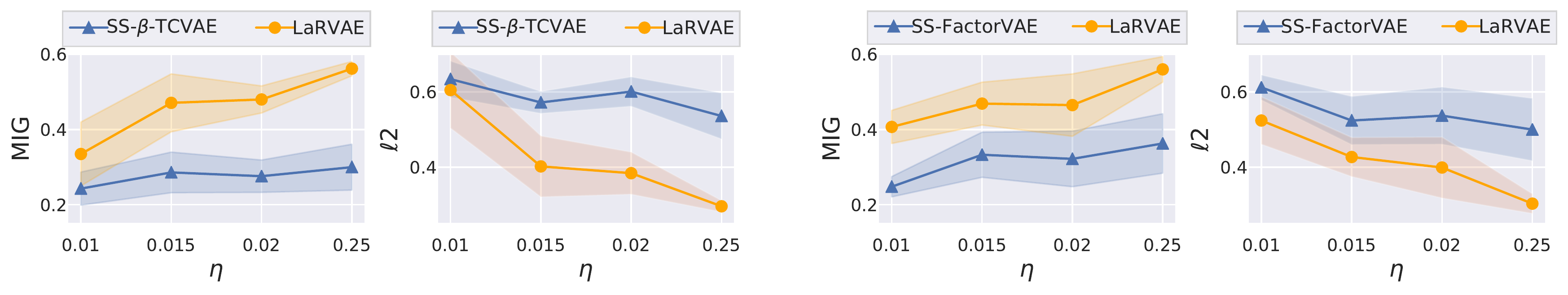}}
		\vspace{-6mm}
		\caption{\small dataset: dSprites}
  \end{subfigure}\\
  \vspace{10pt}
  \begin{subfigure}[b]{1\textwidth}
		\centering
		\small
		\makebox[\textwidth][c]{\includegraphics[width=1.025\linewidth]{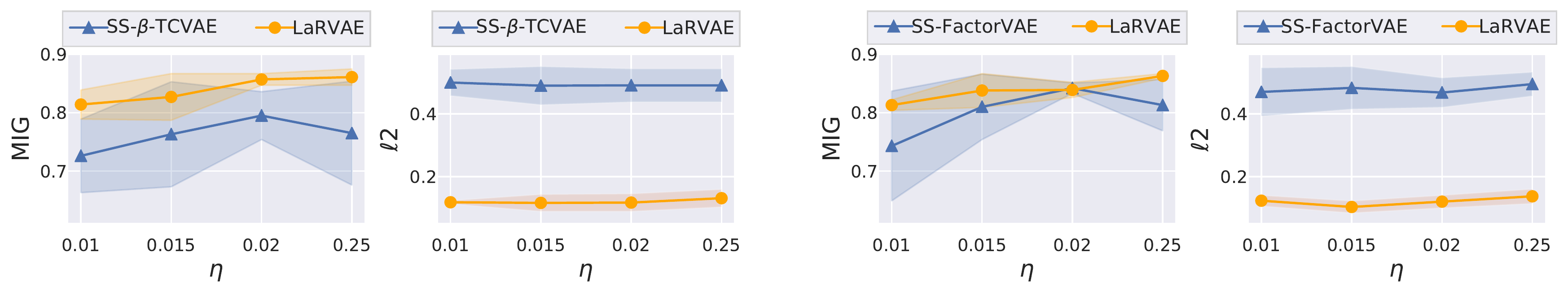}}
		\vspace{-6mm}
		\caption{\small dataset: 3DShapes}
  \end{subfigure}\\
  \vspace{10pt}
  \begin{subfigure}[b]{\textwidth}
		\centering
		\small
		\makebox[\textwidth][c]{\includegraphics[width=1.025\linewidth]{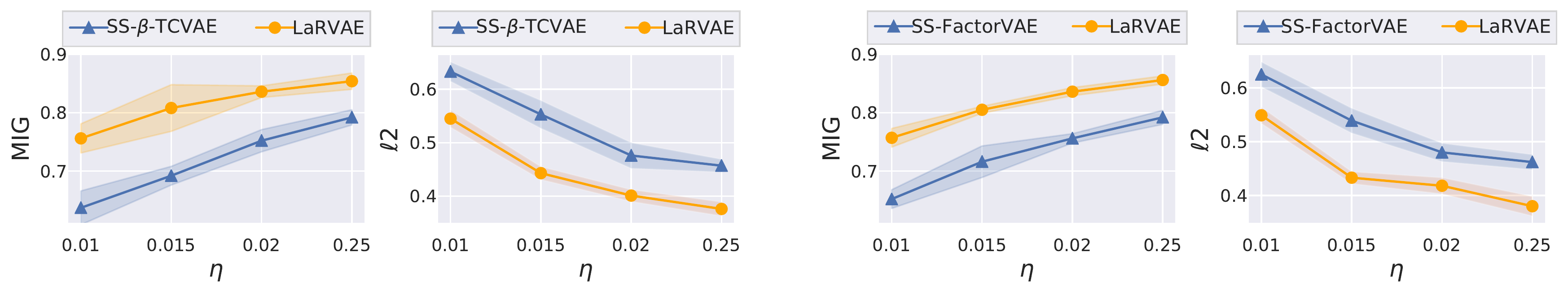}}
		\vspace{-6mm}
		\caption{\small dataset: Isaac3D }
  \end{subfigure}
  \caption{\small Disentanglement performances comparing LaRVAE to 2 semi-supervised baselines (SS-$\beta$-TCVAE and SS-FactorVAE) on 4 different label rates ($\eta$= \{1\%, 1.5\%, 2\%, 2.5\%\}), 3 datasets (dSprites, 3DShepes and Isaac3D) and 2 metrics (MIG and $\ell_2$). In most cases, LaRVAE significantly outperforms baselines.}
  \label{ss}
  \vspace{-5pt}
\end{figure}

{\bf    datasets.}~  
We use 3 synthetic datasets ---  dSprites~\cite{dsprites17}, 3DShapes~\cite{3dshapes18} and Isaac3D~\cite{nie2020semi} ---
which are standard test cases for disentanglement learning.
These synthetic datasets include the fully observed ground-truth labels, which enable comprehensive quantitative evaluations using disentanglement metrics.
During training, we only sample $\eta$ of all ground-truth labels available to the model in order to simulate the semi-supervised setting. In this experiment, $\eta=\{0.01, 0.015, 0.02, 0.025\}$.

{\bf    Evaluation metrics.} 
We use the mutual information gap (MIG)~\cite{betatcvae} to evaluate disentanglement. FactorVAE score is also considered but we present it in the Supplementary Material because it correlate well with MIG~\citep{locatello2019challenging}.
We also use the $\ell_2$ score, i.e., the Euclidean distance between the inferred and the ground-truth labels, to measure the semantic correctness of the inferred label.  
For the MIG score, higher is better; for the $\ell_2$ score, lower is better. 

{\bf    Baselines.} 
We consider 3 semi-supervised disentanglement VAEs as baselines including SS-$\beta$-VAE, SS-$\beta$-TCVAE and SS-FactorVAE~\cite{locatello2019-semi-disentangling}.
These baselines add a label loss $\mathcal{L}_{\rm recon}$ to their unsupervised counterparts and thus differ only in their implementation of the unsupervised regularization (2nd term in Eq.~\ref{unsup}).
We implement LaRVAE with these baselines, which practically adds a label replacement loss $\mathcal{L}_{\rm rep}$ to each baseline.
We then compare each baseline with its LaRVAE version.
We train all models for 1 million iterations using the Adam~\citep{adam} optimizer with a constant learning rate of $0.0001$. Each experiment is repeated 6 times with random seeds.

{\bf Results.} Figure~\ref{ss} presents the quantitative evaluation results.
Comparison between LaRVAE and SS-$\beta$-VAE shows similar trends and is deferred to the Supplementary Material.
%
Figure~\ref{ss} clearly demonstrates that LaRVAE outperforms baselines for both the MIG and $\ell_2$ metrics across all 3 datasets. For many cases, the improvement is statistically significant where the performance difference between LaRVAE and baseline is beyond one level of standard deviation. We further note that, in general, LaRVAE has less variance in its performance and that the variance reduces as more ground-truth labels are available. These observations suggest LaRVAE's disentanglement learning performance is more stable and consistent than the baselines.

\subsection{Effects of Hyperparameters}
We investigate the effects of 2 hyperparameters including $\tau$ that controls the strength of $\mathcal{L}_{\rm rep}$ and the dimension of the nuisance $\rvz$.
We conduct experiments on the dSprites dataset and on two label rates $\eta=\{0,01, 0.02\}$, using SS-$\beta$-TCVAE to implement LaRVAE. 

\begin{figure}
  \centering
  \begin{subfigure}[b]{0.48\textwidth}
		\centering
		\small
		\includegraphics[width=\linewidth]{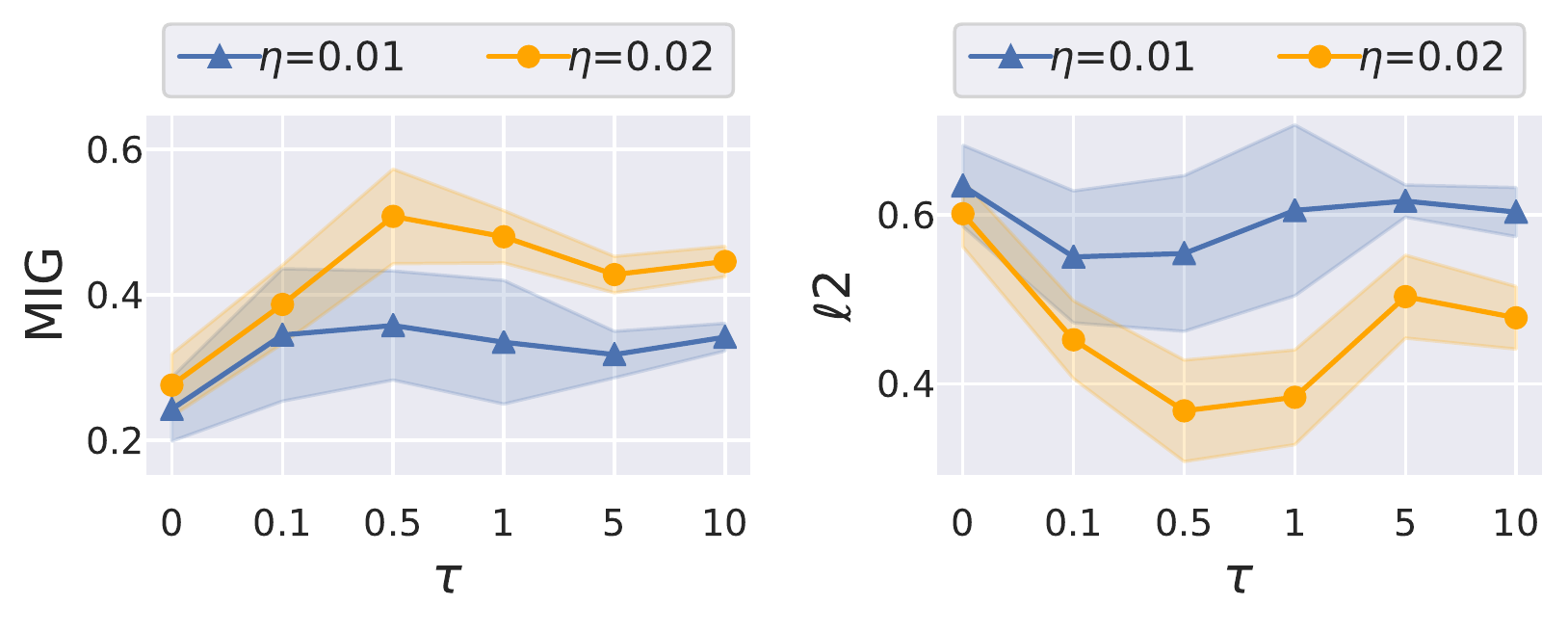}
		\vspace{-5mm}
		\caption{Varying $\tau$.}
		\label{fig:tuning-alpha}
  \end{subfigure}
\quad
  \begin{subfigure}[b]{0.48\textwidth}
		\centering
		\small
		\includegraphics[width=\linewidth]{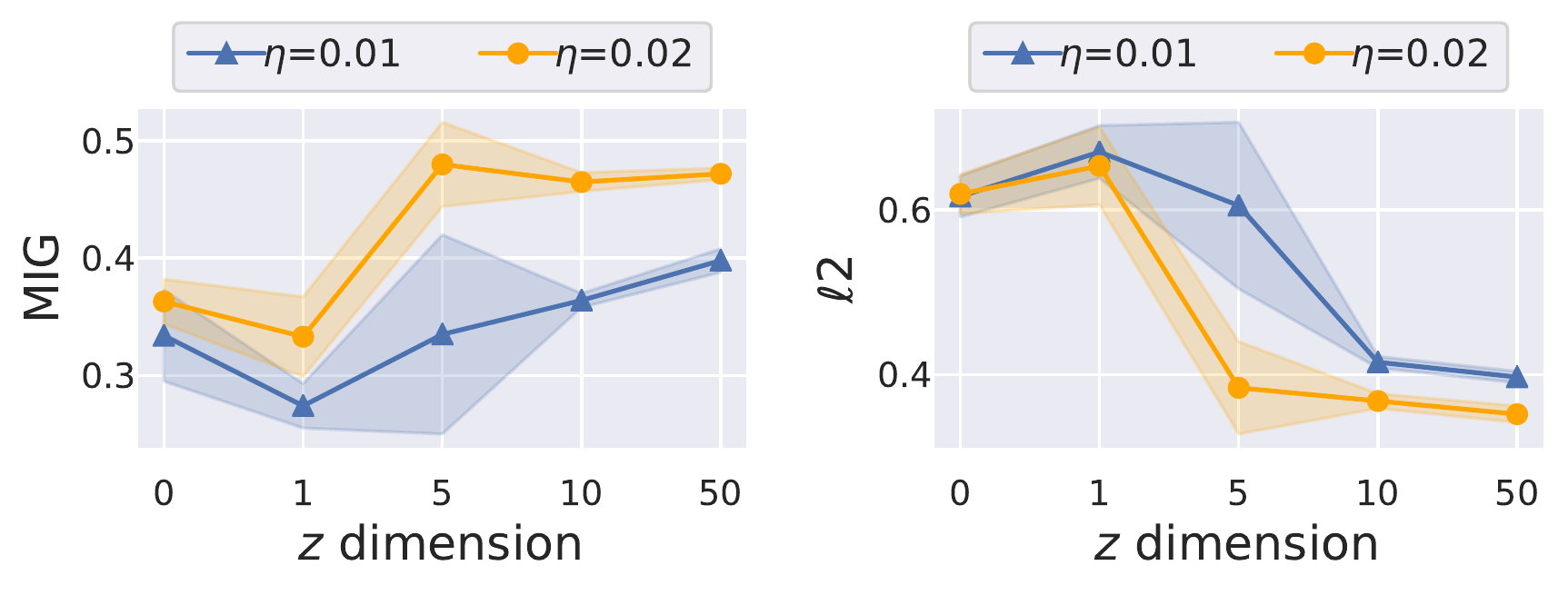}
		\vspace{-5mm}
		\caption{Varying $\rvz$ dimension.}
		\label{fig:tuning-z}
  \end{subfigure}
  \caption{\small Effects of hyperparameters including strength of the $\mathcal{L}_{\rm rep}$ regularization (Figure~\ref{fig:tuning-alpha}) and the  dimension of the nuisance $\rvz$ (Figure~\ref{fig:tuning-z}) on the dSprites dataset. Figure~\ref{fig:tuning-alpha} suggests that $\mathcal{L}_{\rm rep}$ needs to be tuned for different number of available labels while Figure~\ref{fig:tuning-z} suggests that a larger $\rvz$ seems to always lead to better results.}
  \label{hyper_tune}
  \vspace{-5pt}
\end{figure}

{\bf    Strength of the label replacement regularization $\mathcal{L}_{\rm rep}$.}~
Figure~\ref{fig:tuning-alpha} reports MIG and $\ell_2$ scores with varying $\tau=\{0, 0.1, 0.5, 1, 5, 10\}$. %
We can observe that disentanglement performance improves for $\tau > 0$, which again demonstrates the benefit of using the label replacement loss $\mathcal{L}_{\rm rep}$. 
We can also observe a typical regularization effect, i.e., the disentanglement performance first improves then drops with increasing $\tau$. Figure~\ref{fig:tuning-alpha} implies a trade-off exists between the strength of the regularization and the disentanglement performance and suggests that $\tau$ needs to be tuned for different datasets and for different label rates to achieve optimal disentanglement.

{\bf    Dimension of the nuisance $\rvz$.}
In principle, if the label $\rvy$ fully captures all factors of variation of a given data point, then we can omit the nuisance $\rvz$, because it adds no information.
However,
rarely in reality do we have complete knowledge of all of the factors of variation of a data point. 
In such a situation, only a part of factors of variation is observed in the label $\rvy$, and it is desirable to include $\rvz$ to represent factors of variation not captured by $\rvy$.
Even though our work considers fully observed $\rvy$, it is interesting to observe how the disentanglement performance changes when the latent variable dimension is larger than the label dimension. The hypothesis is that the larger freedom in the latent space may stabilize the VAE training. Besides, this provides insight into how LaRVAE can be modified for the partially observed label setting~\cite{Shu2020Weakly, 2020arXiv200202886L}, which we leave as the future work.
%

Figure~\ref{fig:tuning-z} reports the disentanglement performance with varying $\rvz$ dimensions chosen from $\{0, 1, 5, 10, 50\}$. Interestingly, we observe that increasing the dimension of $\rvz$ seems to always improve disentanglement for large enough $\rvz$ (dimension $\geq 5$). This suggests that, using LaRVAE, having a large $\rvz$ does not cause information to leak from $\rvy$ to $\rvz$ as one would expect. Thus, one may wish to use a large $\rvz$ in LaRVAE in practice to achieve better disentanglement performance. 

\subsection{Label Traversal Visualizations}
{\bf Setup.} We perform a label traversal experiment to visually demonstrate the superior disentanglement that LaRVAE learns compared to the baselines. 
Note that, for this experiment, we have access to the label $\rvy$ of each data point $\rvx$ in a given dataset, which enables us to compare the images generated from the models using $\rvy$ as input with the reference, ground-truth image $\rvx$ in the dataset corresponding to $\rvy$.
We first randomly select a label from the dataset.
For each dimension 
of a chosen label $\rvy$, we vary its value while keeping the other dimensions fixed
, i.e., $\widetilde{\rvy}(k,c) = \rvy|_{y_k = c}$ where $c\in [{\rm min}(y_k), {\rm max}(y_k)]$.
We then feed 
$\widetilde{\rvy}$'s 
to LaRVAE and the baseline implemented with SS-$\beta$-TCVAE. 
Importantly, the reference image $\rvx$ is not needed as input because our setup assumes the labels have fully captured all factors of variation. 
For this experiment, we show results on 3DShapes for best visual demonstration. We additionally train and evaluate models on CelebA~\cite{liu2015faceattributes} to demonstrate traversal on real-world dataset.
Many more traversal examples on the remaining 2 datasets dSprites and Isaac3D are available in the Supplementary Material.

{\bf Results.} Figure~\ref{fig:traveral-3dshapes} visualizes the traversal results for the 3DShapes (Figure 4a-c) and the CelebA (Figure 4d-f) datasets, each for 3 selected label dimensions (for 3DShapes: object color, object size, and object shape; for CelebA: pale face, bangs, and glasses). 
The leftmost image in each sub-figure is the reference image $\rvx$ corresponding to the selected label to be varied. The 5 right images in the top and bottom rows in each sub-figure corresponds to the images generated from the baseline (SS-$\beta$-TCVAE) and LaRVAE, respectively. 

We make 2 important observations. First, in most cases, for the label dimension that is varied, LaRVAE controllably generates images with the corresponding attribute varied, whereas baselines fail to do so. For example, in the middle plot in Figure~\ref{fig:traversal-3dshapes-2}, LaRVAE successfully generates images with varying colors, whereas the baseline generated images do not change color but rather shape, which does not correspond to the label dimension that is varied.
Second, in most cases LaRVAE generates images with attributes correctly specified by the label dimensions that are fixed, whereas baseline fails to do so. 
For example, in the middle plot in Figure~\ref{fig:traversal-3dshapes-3}, all colors in the LaRVAE generated images are the same as the images corresponding to the label, whereas the baseline generated images contain wrong colors for the object and the wall.

We note that sometimes there is mismatch between certain attributes in the LaRVAE generated images and the ground-truth image corresponding to a selected label. 
For example, LaRVAE sometimes generates images with incorrect 
colors (e.g., the wall color of the images in the bottom row in Figure~\ref{fig:traversal-3dshapes-3}), suggesting room for improvement. 
Nevertheless, the above experimental results clearly demonstrate that LaRVAE outperforms the baselines both quantitatively and qualitatively on disentanglement learning.

\begin{figure*} [t]
    \centering
    \begin{subfigure}{0.32\linewidth}
    \centering
    \includegraphics[width=\linewidth]{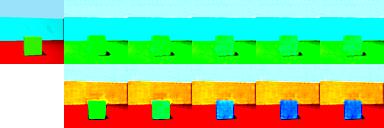}\\[3pt]
    \includegraphics[width=\linewidth]{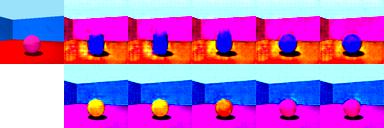}\\[3pt]
    \includegraphics[width=\linewidth]{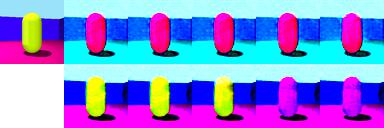}
    \subcaption{3DShapes: Object color}
    \label{fig:traversal-3dshapes-2}
    \end{subfigure}
    \begin{subfigure}{0.32\linewidth}
    \centering
    \includegraphics[width=\linewidth]{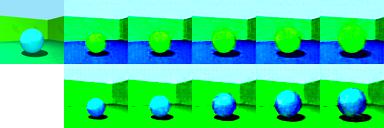}\\[3pt]
    \includegraphics[width=\linewidth]{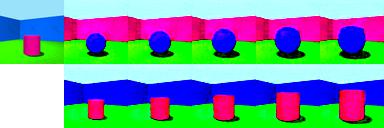}\\[3pt]
    \includegraphics[width=\linewidth]{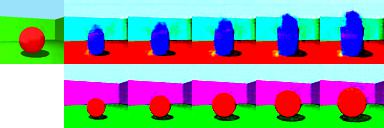}
    \subcaption{3DShapes: Object size}
    \label{fig:traversal-3dshapes-3}
    \end{subfigure}
    \begin{subfigure}{0.32\linewidth}
    \centering
    \includegraphics[width=\linewidth]{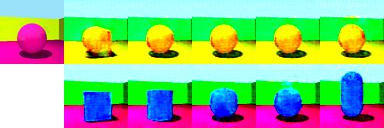}\\[3pt]
    \includegraphics[width=\linewidth]{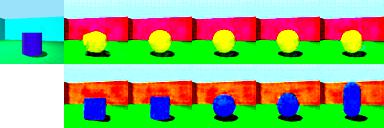}\\[3pt]
    \includegraphics[width=\linewidth]{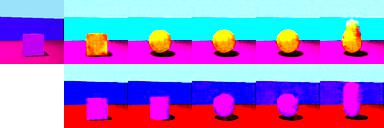}
    \subcaption{3DShapes: Object shape}
    \label{fig:traversal-3dshapes-4}
    \end{subfigure}
    \\[7.5pt]    
    \begin{subfigure}{0.32\linewidth}
    \centering
    \includegraphics[width=\linewidth]{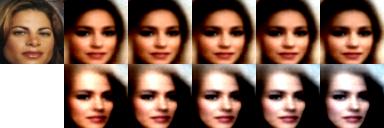}\\[3pt]
    \includegraphics[width=\linewidth]{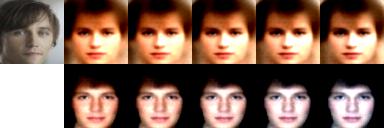}\\[3pt]
    \includegraphics[width=\linewidth]{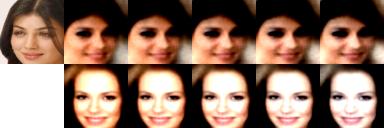}
    \subcaption{CelebA: pale face}
    \label{fig:traversal-celeba-pale-face}
    \end{subfigure}
    \begin{subfigure}{0.32\linewidth}
    \centering
    \includegraphics[width=\linewidth]{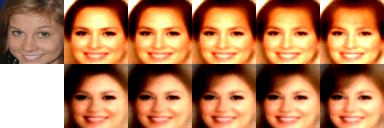}\\[3pt]
    \includegraphics[width=\linewidth]{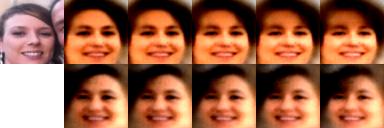}\\[3pt]
    \includegraphics[width=\linewidth]{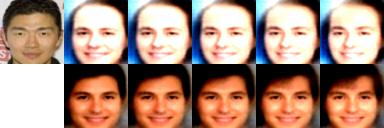}
    \subcaption{CelebA: bangs}
    \label{fig:traversal-celeba-bangs}
    \end{subfigure}
    \begin{subfigure}{0.32\linewidth}
    \centering
    \includegraphics[width=\linewidth]{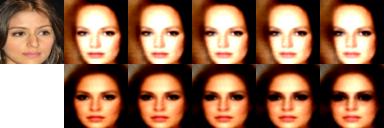}\\[3pt]
    \includegraphics[width=\linewidth]{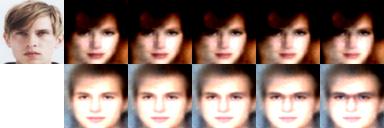}\\[3pt]
    \includegraphics[width=\linewidth]{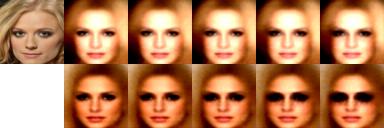}
    \subcaption{CelebA: glasses}
    \label{fig:traversal-celeba-glasses}
    \end{subfigure}
    \caption{\small
    Generated images by label traversal comparing LaRVAE (bottom row in each plot) and SS-$\beta$-TCVAE baseline (top row, column 1-6 in each plot) on the 3DShapes dataset (Figure~\ref{fig:traversal-3dshapes-2}-\ref{fig:traversal-3dshapes-4}) and the CelebA dataset (Figure~\ref{fig:traversal-celeba-bangs}-\ref{fig:traversal-celeba-glasses}). 
    The leftmost column in each plot is the reference image.
    LaRVAE clearly disentangles the selected dimensions of the label better than the baseline while, most visibly for the 3DShapes dataset, keeps the attributes in the generated image the same as specified by the remaining label dimensions.
    }
    \label{fig:traveral-3dshapes}
    \vspace{-7.5pt}
\end{figure*}

\vspace{-2.5pt}
\section{Conclusions}
\vspace{-2.5pt}
In this work, we have studied semi-supervised disentanglement learning under the VAE framework. We build on and extend~\cite{locatello2019-semi-disentangling} by introducing a label replacement regularization which substitutes the inferred label with the true label, whenever it is available, during training. 
We have also shown how our general formulation of semi-supervised disentanglement learning under the VAE setting introduces both the label replacement and the label loss terms in a principle manner. 
Quantitative and qualitative experimental results on both synthetic and real datasets demonstrate the superior disentanglement performance of our extension compared to the baselines in~\cite{locatello2019-semi-disentangling}. 
The promising results in the present work encourages other ways to effectively exploiting information in the semi-supervised setting to further improvements disentanglement learning.
For example, recent progress in semi-supervised learning methods that innovatively leverage labeled and unlabeled data~\cite{berthelot2019mixmatch,izmailov2019semi,zhai2019s4l} are especially inspiring, which could motivate new methodologies for disentanglement learning with limited supervision. 

\section*{Acknowledgements}
WN and ABP were supported by IARPA via DoI/IBC contract D16PC00003.
ZW and RGB were supported by 
NSF grants CCF-1911094, IIS-1838177, and IIS-1730574; 
ONR grants N00014-18-12571 and N00014-17-1-2551;
AFOSR grant FA9550-18-1-0478; 
DARPA grant G001534-7500; and a 
Vannevar Bush Faculty Fellowship, ONR grant N00014-18-1-2047.

{\small

\bibliography{bib}
\bibliographystyle{abbrvnat}}

\newpage


  









\appendix
\counterwithin{table}{section}
\counterwithin{figure}{section}

\renewcommand{\theequation}{\Alph{section}.\arabic{equation}}
\vspace{-20pt}
\section{Deriving the Label Replacement Disentanglement VAE Objective (Eq. 7)}

We start from the generalized negative log-likelihood (NLL) for semi-supervised VAEs, which is
\begin{align} \label{eq:nll_orig}
    \begin{split}
        \mathcal{L}_{\rm semi} = & \mathcal{L}'_{\rm unsup} + \gamma\mathbb{E}_{\rvx, \rvy\sim\mathcal{P}_{L}}[-{\rm log}\,p_\theta(\rvx, \rvy)]
    \end{split}
\end{align} 
where 
\begin{align} \label{eq:unsup}
    \mathcal{L}'_{\rm unsup} = \mathbb{E}_{\rvx \sim \mathcal{P}_{U}} [{-{\rm log}\,p_\theta(\rvx)}] - \gamma_{\rm tc} \mathbb{E}_{\rvx}[R_u(q_{\phi}(\bm{\xi} | \rvx))]
\end{align}
which is a variational upper bound of $\mathcal{L}_{\rm unsup}$ in Eq. 1. 

By decomposing the log of the joint distribution $p_\theta(\rvx, \rvy)$ into
\begin{align} \label{bayes}
    {\rm log}\,p_\theta(\rvx, \rvy) = \lambda {\rm log}\,p_\theta(\rvy|\rvx) p_\theta(\rvx) + (1 - \lambda){\rm log}\, p_\theta(\rvx|\rvy) p_\theta(\rvy)
\end{align}
we have 
\begin{align} \label{L_semi_1}
\begin{split}
    \mathcal{L}_{\rm semi} =  \mathcal{L}'_{\rm unsup} + \gamma \lambda \mathbb{E}_{\rvx, \rvy\sim\mathcal{P}_{L}}[-{\rm log}\,p_\theta(\rvy|\rvx) p_{\theta}(\rvx)] +  \gamma(1- \lambda) \mathbb{E}_{\rvx,\rvy\sim\mathcal{P}_L}[-{\rm log}\,p_\theta(\rvx|\rvy) p_{\theta}(\rvy)] \\
\end{split}
\end{align}
By substituting Eq. \ref{eq:unsup} into Eq. \ref{L_semi_1} and doing some algebraic arrangements, we have 
\begin{align} \label{L_semi_2}
\begin{split}
    \mathcal{L}_{\rm semi} = & (1+\gamma\lambda) \mathbb{E}_{\rvx \sim \mathcal{P}_{U}} [{-{\rm log}\,p_\theta(\rvx)}] - \gamma_{\rm tc} \mathbb{E}_{\rvx}[R_u(q_{\phi}(\bm{\xi} | \rvx))] + \gamma \lambda \mathbb{E}_{\rvx, \rvy\sim\mathcal{P}_{L}}[-{\rm log}\,p_\theta(\rvy|\rvx)] \\ & +  \gamma(1- \lambda)  \mathbb{E}_{\rvx,\rvy\sim\mathcal{P}_L}[-{\rm log}\,p_\theta(\rvx|\rvy) - \log p_{\theta}(\rvy)] \\
    \mathop  = \limits^{\left( a \right)} & \mathbb{E}_{\rvx \sim \mathcal{P}_{U}} [{-{\rm log}\,p_\theta(\rvx)}] - \gamma'_{\rm tc} \mathbb{E}_{\rvx}[R_u(q_{\phi}(\bm{\xi} | \rvx))] + \alpha \mathbb{E}_{\rvx, \rvy\sim\mathcal{P}_{L}}[-{\rm log}\,p_\theta(\rvy|\rvx)] \\ & +  \tau \mathbb{E}_{\rvx,\rvy\sim\mathcal{P}_L}[-{\rm log}\,p_\theta(\rvx|\rvy) - \log p_{\theta}(\rvy)] \\
    \mathop  = \limits^{\left( b \right)} & \mathcal{L}'_{\rm unsup} + \alpha \mathbb{E}_{\rvx, \rvy\sim\mathcal{P}_{L}}[-{\rm log}\,p_\theta(\rvy|\rvx)] +  \tau \mathbb{E}_{\rvx,\rvy\sim\mathcal{P}_L}[-{\rm log}\,p_\theta(\rvx|\rvy)] \\
    \mathop  \approx \limits^{\left( c \right)} & \mathcal{L}_{\rm unsup} + \alpha \mathbb{E}_{\rvx, \rvy\sim\mathcal{P}_{L}}[-{\rm log}\,q_\phi(\rvy|\rvx)] +  \tau \mathbb{E}_{\rvx,\rvy\sim\mathcal{P}_L}[-{\rm log}\,p_\theta(\rvx|\rvy)]
\end{split}
\end{align}
where $(a)$ is from dividing two sides by a constant coefficient $(1+\gamma\lambda)$, and setting $\gamma'_{\rm tc} = \frac{\gamma_{\rm tc}}{1 + \lambda \gamma}$, $\alpha = \frac{\lambda \gamma}{1 + \lambda \gamma}$ and $\tau = \frac{ (1-\lambda) \gamma}{1 + \lambda \gamma}$. $(b)$ follows from the definition of $\mathcal{L}'_{\rm unsup}$ in Eq. \ref{eq:unsup} ($\gamma'_{\rm tc}$ and $\gamma'_{\rm tc}$ are interchangeable as they are tunable hyparameters), and also from the fact that in VAEs, the prior $p_\theta(\rvx|\rvy)$ is usually assumed to be a standard Gaussian and thus does not involve any model parameters. Finally, $(c)$ is from the fact that $\mathcal{L}'_{\rm unsup}$ can be approximated by $\mathcal{L}_{\rm unsup}$ in Eq. 1, and the fact that the posterior $p_\theta(\rvy|\rvx)$ is usually intractable in VAEs, and thus we use another parametrized variational distribution $q_\phi(\rvy|\rvx)$ to approximate it.

\section{Deriving the Supervised Regularizations (Eqs. 8 and 9)}

First, we know 
\begin{align} \label{L_recon}
    \mathcal{L}_{\rm recon} = \mathbb{E}_{\rvx, \rvy\sim\mathcal{P}_{L}}[-{\rm log}\,q_\phi(\rvy|\rvx)]
\end{align}
and the approximate posterior is parametrized as 
\begin{align} \label{gaussian_y_x}
    {q}_{\phi} (\rvy | \rvx) = \mathcal{N} ({\bm{\mu}}_{y_{\phi}}(\rvx), \sigma^2 \mI )
\end{align}
By plugging Eq. \ref{gaussian_y_x} into Eq. \ref{L_recon}, we have 
\begin{align} \label{L_recon_v2}
    \mathcal{L}_{\rm recon} \propto \mathbb{E}_{\rvx, \rvy\sim\mathcal{P}_{L}}[\frac{1}{2\sigma^2} \| {\bm{\mu}}_{y_{\phi}}(\rvx) - \rvy \|^2 ]
\end{align}
By setting $\alpha=\frac{1}{2\sigma^2}$ and neglecting the constant proportional coefficient, we obtain Eq. 8.

Second, we know 
\begin{align}
    \mathcal{L}_{\rm rep} = \mathbb{E}_{\rvx,\rvy\sim\mathcal{P}_L}[-{\rm log}\,p_\theta(\rvx|\rvy)]
\end{align}
We then evaluate $\mathcal{L}_{\rm rep}$ with its average ELBO as follows,
\begin{align} \label{cp}
    \begin{split}
        \mathcal{L}_{\rm rep} &= -\mathbb{E}_{\rvx,\rvy\sim\mathcal{P}_L} \left[{\rm log}\int_{z}p_\theta(\rvx|\rvy, \rvz)p(\rvz)d\rvz \right] \\
    &\leq -\mathbb{E}_{\rvx,\rvy\sim\mathcal{P}_L, \rvz\sim q_\phi(\rvz|\rvx)}\left[{\rm log} \frac{p_\theta(\rvx|\rvy, \rvz)p(\rvz)}{q_\phi(\rvz|\rvx)} \right] \\
    &= -\mathbb{E}_{\rvx,\rvy\sim\mathcal{P}_L, \rvz\sim q_\phi(\rvz|\rvx)} \left[{\rm log}\,p_\theta(\rvx|\rvy,\rvz)\right] - \mathbb{E}_{\rvx\sim\mathcal{P}_L} [ D_{\rm KL}(q_\phi(\rvz|\rvx) || p(\rvz))] 
    \end{split}
\end{align}
where the inequality comes from the  Jensen's inequality. As similar to normal VAEs, we assume the likelihood $p_\theta(\rvx|\rvy,\rvz)$ is a parameterized Gaussian for tractability, i.e., $p_{\theta} (\rvx | \bm{\xi}) = \mathcal{N} ({\bm {\mu}}_{\theta}(\bm{\xi}), \sigma^2 \mI )$, and also assume the approximate posterior as ${q}_{\phi} (\rvy | \rvx) = \mathcal{N} ({\bm{\mu}}_{z_{\phi}}(\rvx), \sigma^2 \mI )$.

\section{Training Procedure}

\begin{algo}[H]
\SetKwInOut{Input}{input}\SetKwInOut{Output}{output}
\SetAlgoLined
\Input{Data set $\mathcal{D} = \mathcal{P}_L \bigcup \mathcal{P}_U $, labeled data set $\mathcal{P}_L$, batch size $B$, optimizer \texttt{optim}, learning rate $\eta$, number of iterations $T$, encoder parameter $\phi$, decoder parameter $\theta$}
\Output{Trained parameters $\phi$ and $\theta$}
 initialize $\phi$ and $\theta$\;
 \While{iteration less than $T$}{
  sample batch $\mathcal{B}$ of $\rvx$'s of size $B$ from $\mathcal{D}$;\\
  sample batch $\mathcal{B}_L$ of $(\rvx, \rvy)$'s of size $B$ from $\mathcal{P}_L$;\\
  compute $\mathcal{L}_{\rm unsup}$ on $\mathcal{B}$ (Eq. 1);\\
  compute $\mathcal{L}_{\rm recon}$ and $\mathcal{L}_{\rm rep}$ on $\mathcal{B}_L$ (Eqs. 8 and 9);\\
  compute $\mathcal{L}_{\rm semi}$ (Eq. 7);\\
  update $\phi, \theta := \texttt{optim}(\phi, \theta, \eta, \mathcal{L}_{\rm semi})$;

 }
 \caption{Training Procedure}
\end{algo}

\section{Additional Experiment Setup}

\paragraph{Dataset.} Table~\ref{tab:dataset} summarizes the basic statistics of all 4 datasets used in our experiments. Note that \#Factors = dimension of label $\rvy$ and \#latent = dimension of label $\rvy$ + dimension of nuisance $\rvz$. For the CelebA dataset, we additionally crop and align so each image contain only the face without background and resize to 64$\times$64 by simple downsampling. For the Isaac3D dataset, we resize each image to 64$\times$64 with a bilinear downsampling.
\begin{table}[t]
\small
\centering
\caption{\small Summary statistics of the experimental datasets.}
\begin{tabular}{lccc}
\toprule
\textbf{dataset} & \textbf{\#Images} & \textbf{Image size} & \textbf{\#Factors} \\ \hline
dSprites~\cite{dsprites17}         & 737,280            & 64x64x1             & 5                          \\
3DShapes~\cite{3dshapes18}        & 480,000            & 64x64x3             & 6                          \\
Isaac3D~\cite{nie2020semi}         & 737,280           & 64x64x3              & 9 \\
CelebA~\cite{liu2015faceattributes}           & 202,599            & 64x64x3             & 40                         \\ \bottomrule
\end{tabular}
\label{tab:dataset}
\end{table}

\paragraph{Model Architecture.} Table~\ref{tab:model_architecture} summarizes the encoder and decoder architectures used in both LaRVAE and baselines.

\paragraph{Hyperparameters.} Most of the hyperparameter configurations can be found in the training scripts in the \texttt{scripts} folder in the code for this paper. In particular, we use $\tau=0.05$ for $\mathcal{L}_{\rm rep}$ in the CelebA label traversal experiments and $\tau=1$ for all other experiments except for the hyperparameter tuning experiments in Section 5.2.
We set the dimension size of the nuisance $\rvz$ to be five in all experiments except for the hypermarameter tuning experiments. 

\begin{table*}[t!]
\small
\centering
\caption{\small Encoder and decoder architectures used in all experiments.
\vspace{-5pt}
}
\begin{tabular}{ll}
\toprule
\textbf{Encoder}                               & \textbf{Decoder}                                     \\ \hline
input: image width  $\times$ image height $\times$ \#channels & input: \#latent                                      \\
32 4$\times$4 conv, Instance Norm, ReLU, stride 2     & FC 256, ReLU                                          \\
32 4$\times$4 conv, Instance Norm, ReLU, stride 2     & FC 1024, ReLU                                        \\
64 2$\times$2 conv, instance norm, ReLU, stride 2     & 64 4$\times$4 transpose conv, Instance Norm, ReLU, stride 2  \\
64 2$\times$2 conv, instance norm, ReLU, stride 2     & 32 4$\times$4 transpose conv, Instance Norm, ReLU, stride 2 \\
FC 256, FC 2$\times$\#latent                          & 32 4$\times$4 transpose conv, Instance Norm, ReLU, stride 2 \\
                                               & \#channels 4$\times$4 transpose conv, stride 2              \\ \bottomrule
\end{tabular}
\label{tab:model_architecture}
\end{table*}

\paragraph{Label Traversal Setup.} For all label traversal experiments, both the baseline (SS-$\beta$-TCVAE) and LaRVAE (implemented using $\beta$-TCVAE) are trained on 1\% ($\eta$=0.01) of all available labels. 
\paragraph{Hardware.} We mainly use Nvidia V100 (and some RTX2080) GPUs for training. Each GPU can fit multiple experiments because the largest GPU memory taken by our experiments is less than 2GB. Training each model with 1 million iterations takes less than 30 hours at the longest. We only use single GPU for each experiment.

\section{Additional Experimental Results}

\begin{figure} [h]
  \centering
  \begin{subfigure}[b]{0.32\textwidth}
		\centering
		\small
		\makebox[\textwidth][c]{\includegraphics[width=\linewidth]{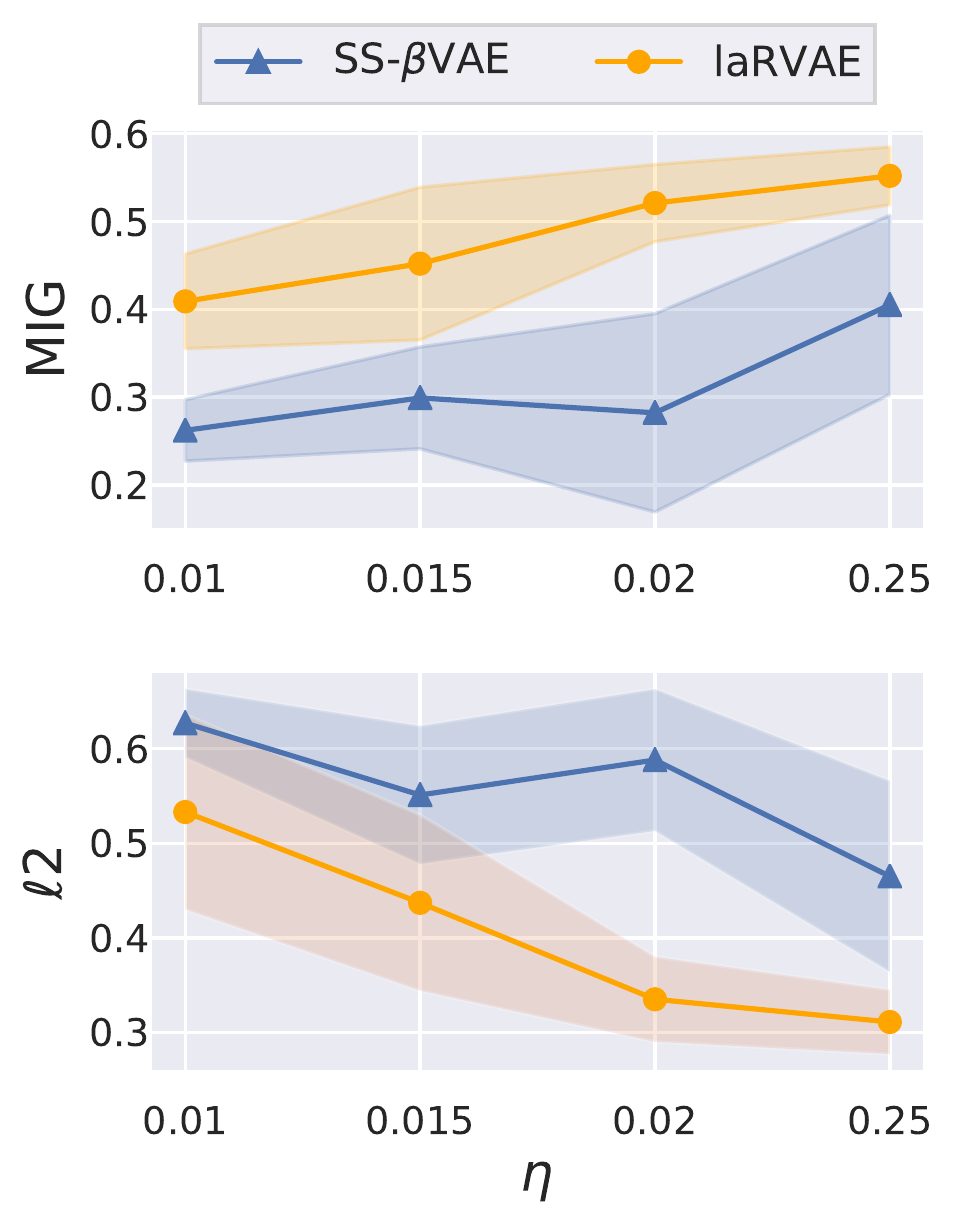}}
		\caption{\small dSprites}
  \end{subfigure}  
  \begin{subfigure}[b]{0.32\textwidth}
		\centering
		\small
		\makebox[\textwidth][c]{\includegraphics[width=\linewidth]{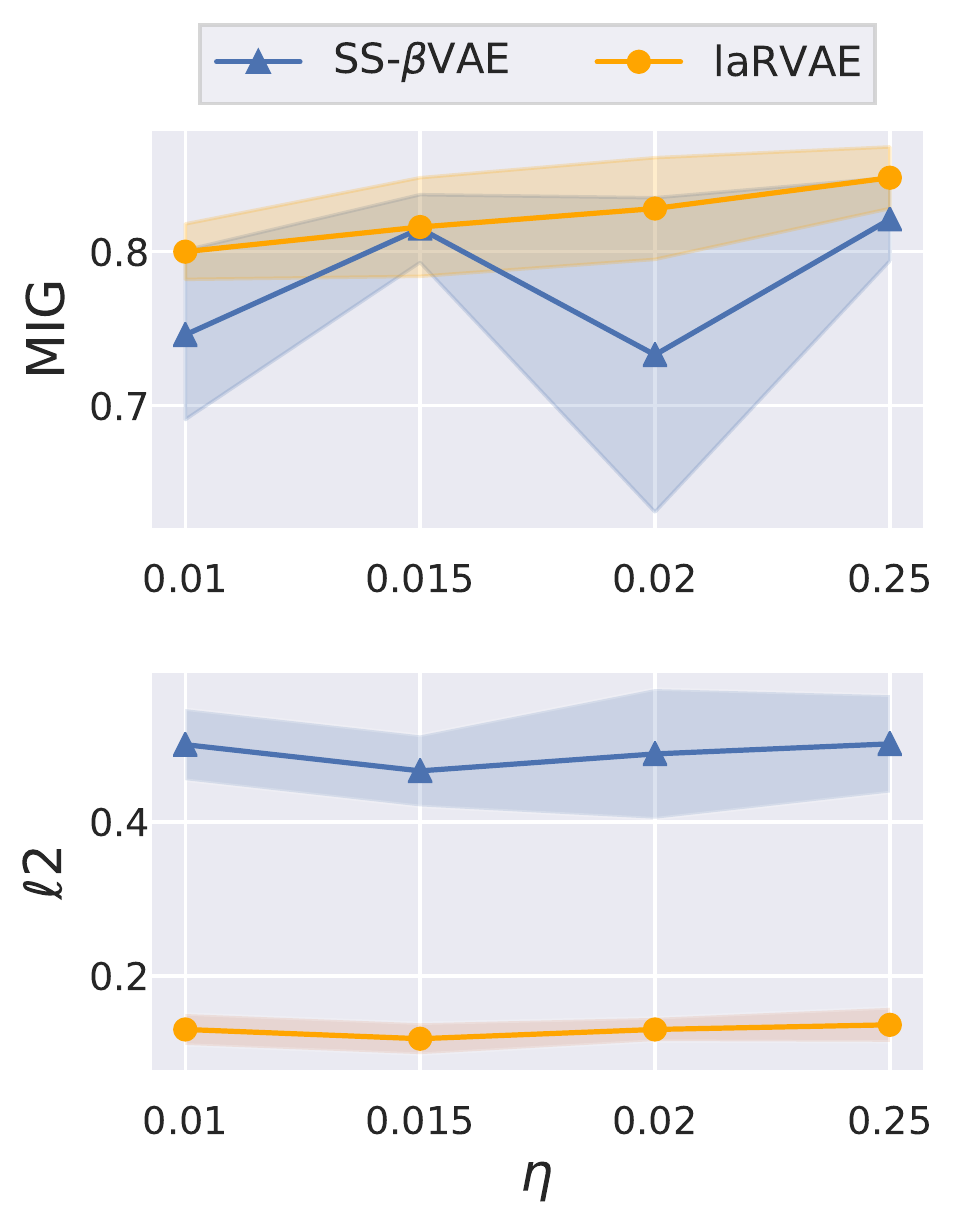}}
		\caption{\small 3DShapes}
  \end{subfigure} 
  \begin{subfigure}[b]{0.32\textwidth}
		\centering
		\small
		\makebox[\textwidth][c]{\includegraphics[width=\linewidth]{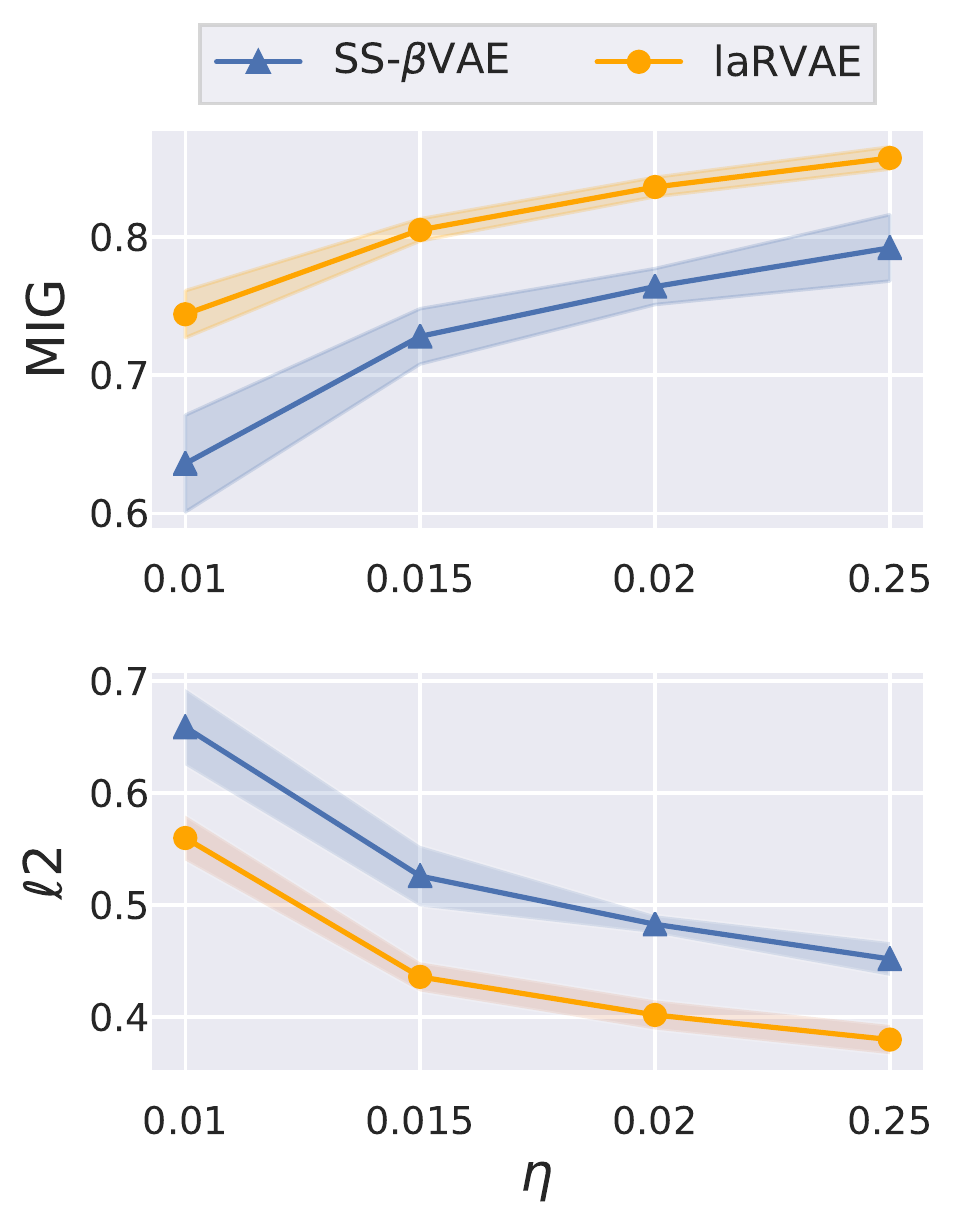}}
		\caption{\small Isaac3D }
  \end{subfigure}
  \caption{\small Disentanglement performances comparing LaRVAE to SS-$\beta$VAE baseline on 4 different label rates ($\eta$= \{1\%, 1.5\%, 2\%, 2.5\%\}), 3 datasets (dSprites, 3DShepes and Isaac3D) and 2 metrics (MIG and $\ell_2$). Similar to the results in the main text (Figure 2), in most cases, LaRVAE significantly outperforms the baseline.}
  \label{fig:betavae}
\end{figure}

\begin{figure} [h!]
  \centering
  \begin{subfigure}[b]{\textwidth}
		\centering
		\small
		\makebox[\textwidth][c]{\includegraphics[width=\linewidth]{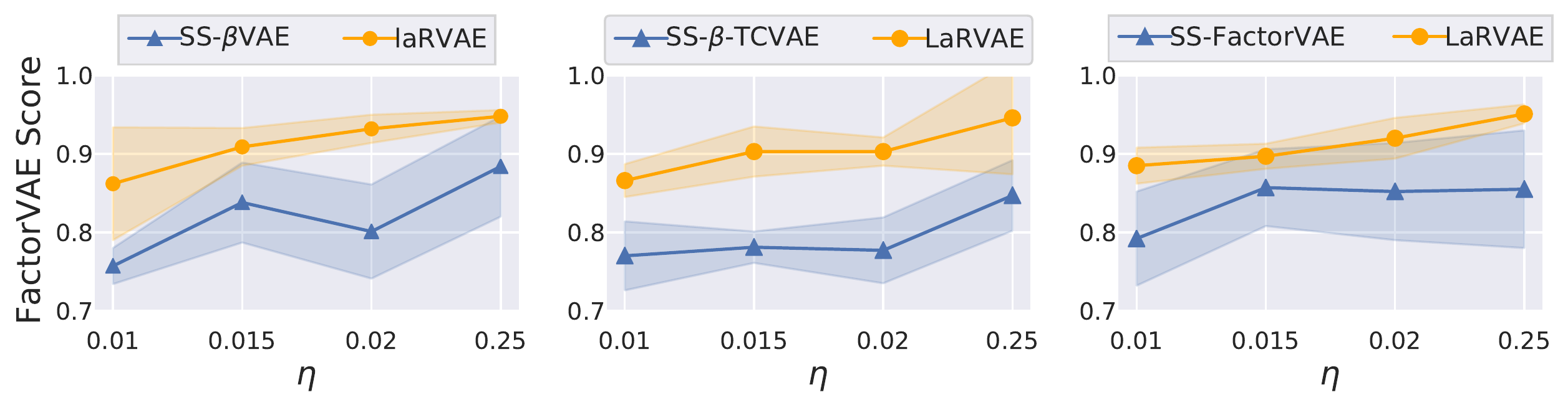}}
		\caption{\small dSprites}
  \end{subfigure}  \\[15pt]
  \begin{subfigure}[b]{\textwidth}
		\centering
		\small
		\makebox[\textwidth][c]{\includegraphics[width=\linewidth]{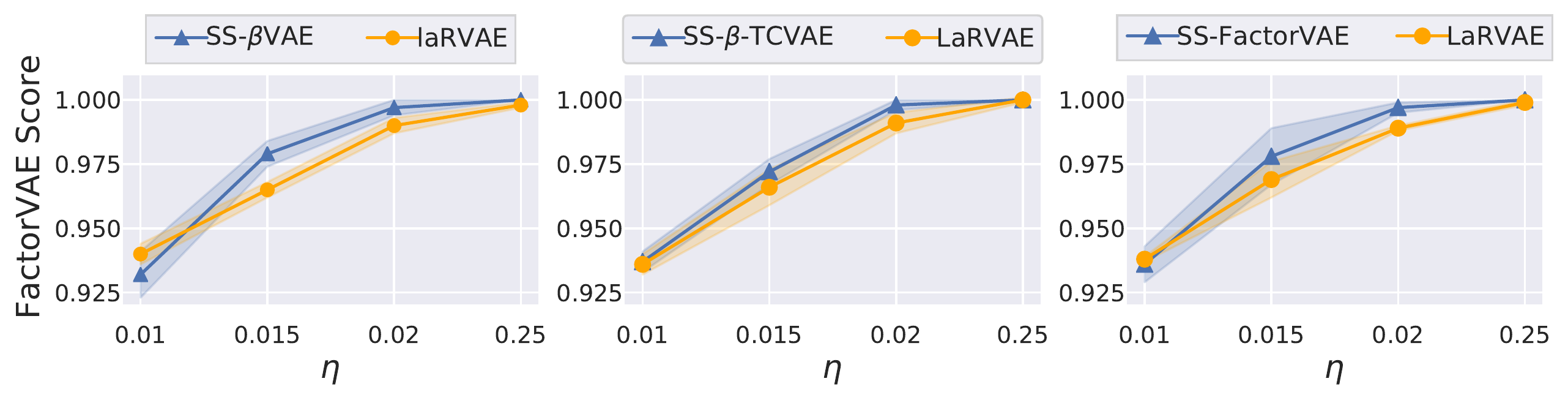}}
		\caption{\small Isaac3D}
  \end{subfigure} 
  \caption{\small FactorVAE score~\cite{factorvae} comparing LaRVAE to each of the three baselines (SS-$\beta$VAE, SS-$\beta$-TCVAE and SS-FactorVAE) on 4 different label rates ($\eta$= \{1\%, 1.5\%, 2\%, 2.5\%\}) and 2 datasets (dSprites and Isaac3D). Similar to the results in the main text (Figure 2), in most cases, LaRVAE significantly outperforms the baseline. Note that the y-axis is very fine grained in Figure E.2b, indicating comparable performance between each baseline and LaRVAE.}
  \label{fig:factorvae}
\end{figure}

\paragraph{Additional Quantitative Evaluation: SS-$\beta$VAE vs. LaRVAE.} 
Figure~\ref{fig:betavae} compares SS-$\beta$VAE with LaRVAE on all three synthetic datasets and on two metrics (MIG and $\ell_2$). We see that, in most cases, LaRVAE significantly improves disentanglement learning upon SS-$\beta$VAE baseline, which is consistent with the findings presented in the main paper.

\paragraph{Additional Quantitative Evaluation: FactorVAE scores.} Figure~\ref{fig:factorvae} presents the FactorVAE score~\cite{factorvae} comparing each of the three baselines (SS-$\beta$VAE, SS-$\beta$-TCVAE and SS-FactorVAE) with its LaRVAE counterpart, respectively, on the dSprites and Isaac3D datasets. 

We did not show results on the 3DShapes dataset because both baselines and LaRVAE achieves perfect FactorVAE score (=1) and thus not meaningful to show. We see that, for the dSprites dataset, LaRVAE obviously improves upon each baseline. For the Isaac3D dataset, LaRVAE achieves FactorVAE score comparable to the baselines (note that the y scale is very fine-grained, showing very close scores).
The fact that both baselines and LaRVAE achieves perfect FactorVAE score on the 3DShapes dataset and almost perfect score on the Isaac3D dataset suggests that 1) the FactorVAE score is close to saturation and there is little room for further improvement in terms of the FactorVAE score and 2) FactorVAE score may not be as an ideal metric for evaluating disentanglement as MIG and $\ell_2$.
Therefore, the results in Figure~\ref{fig:factorvae} are still consistent with the results in the main paper that LaRVAE significantly outperforms baselines in most cases.

\begin{figure} [hp!]
  \centering
  \begin{subfigure}[b]{\textwidth}
		\centering
		\small
		\includegraphics[width=0.32\linewidth]{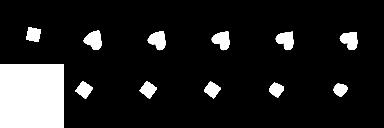}
		\includegraphics[width=0.32\linewidth]{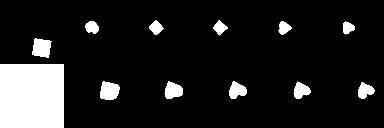}
		\includegraphics[width=0.32\linewidth]{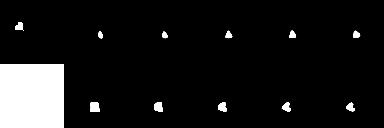}
		\includegraphics[width=0.32\linewidth]{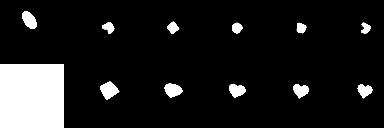}
		\includegraphics[width=0.32\linewidth]{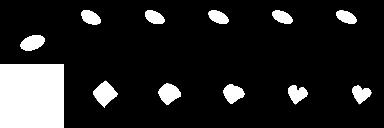}
		\includegraphics[width=0.32\linewidth]{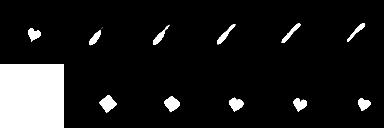}
		\caption{\small dSprites: Object Shape}
  \end{subfigure}  
  \begin{subfigure}[b]{\textwidth}
		\centering
		\small
		\includegraphics[width=0.32\linewidth]{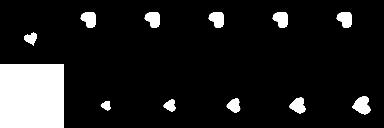}
		\includegraphics[width=0.32\linewidth]{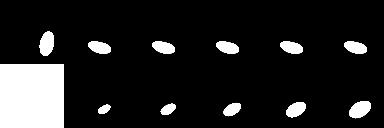}
		\includegraphics[width=0.32\linewidth]{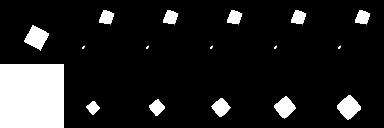}
		\includegraphics[width=0.32\linewidth]{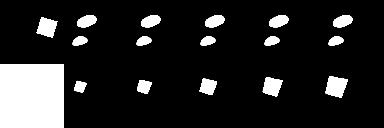}
		\includegraphics[width=0.32\linewidth]{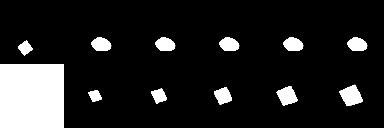}
		\includegraphics[width=0.32\linewidth]{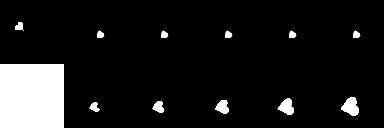}
		\caption{\small dSprites: Object Size}
  \end{subfigure} 
  \begin{subfigure}[b]{\textwidth}
    \centering
		\small
	\includegraphics[width=0.32\linewidth]{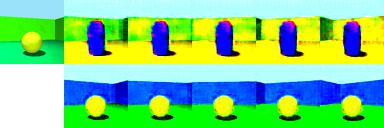}
		\includegraphics[width=0.32\linewidth]{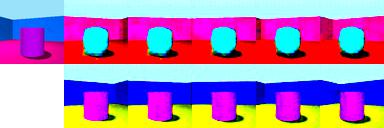}
		\includegraphics[width=0.32\linewidth]{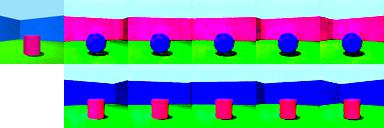}
		\includegraphics[width=0.32\linewidth]{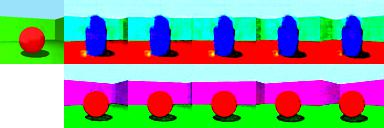}
		\includegraphics[width=0.32\linewidth]{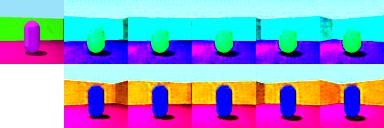}
		\includegraphics[width=0.32\linewidth]{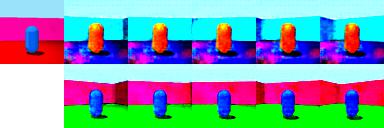}
		\caption{\small 3DShapes: Camera Angle}
  \end{subfigure}
  \begin{subfigure}[b]{\textwidth}
    \centering
		\small
	    \includegraphics[width=0.32\linewidth]{figures/traversal/shapes3d/shapes3d_labels3_38_seed21.jpg}
		\includegraphics[width=0.32\linewidth]{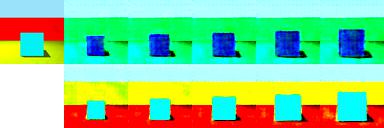}
		\includegraphics[width=0.32\linewidth]{figures/traversal/shapes3d/shapes3d_labels3_24_seed16.jpg}
		\includegraphics[width=0.32\linewidth]{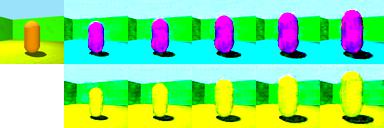}
		\includegraphics[width=0.32\linewidth]{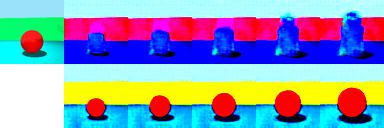}
		\includegraphics[width=0.32\linewidth]{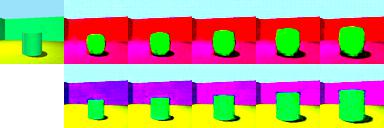}
		\caption{\small 3DShapes: Object Shape}
  \end{subfigure}
   \begin{subfigure}[b]{\textwidth}
    \centering
		\small
	    \includegraphics[width=0.32\linewidth]{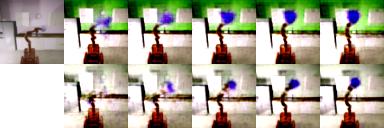}
		\includegraphics[width=0.32\linewidth]{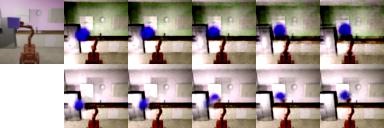}
		\includegraphics[width=0.32\linewidth]{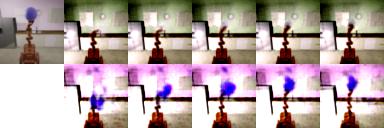}
		\includegraphics[width=0.32\linewidth]{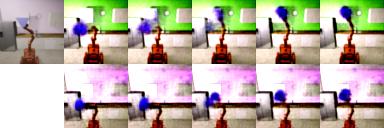}
		\includegraphics[width=0.32\linewidth]{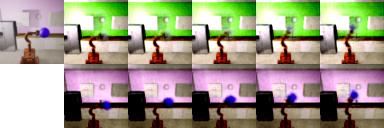}
		\includegraphics[width=0.32\linewidth]{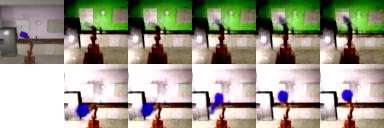}
		\caption{\small Isaac3D: Robot Vertical Movement}
  \end{subfigure}
   \begin{subfigure}[b]{\textwidth}
    \centering
		\small
	    \includegraphics[width=0.32\linewidth]{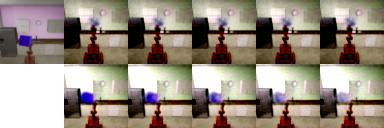}
	    \includegraphics[width=0.32\linewidth]{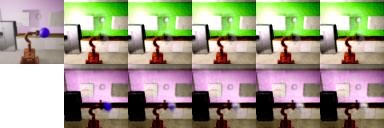}
	    \includegraphics[width=0.32\linewidth]{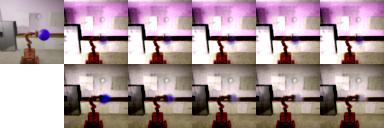}
	    \includegraphics[width=0.32\linewidth]{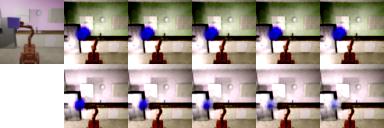}
	    \includegraphics[width=0.32\linewidth]{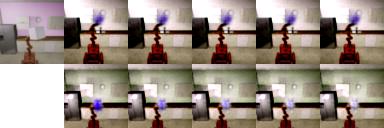}
	    \includegraphics[width=0.32\linewidth]{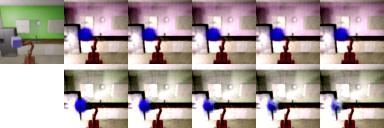}
		\caption{\small Isaac3D: Object Color}
  \end{subfigure}
  \caption{\small 
  Additional label traversal visualizations comparing baseline SS-$\beta$-TCVAE (top row in each plot) and LaRVAE (bottom row in each plot) on selected label dimensions on three datasets (dSprites, 3DShapes and Isaac3D). Leftmost image in each plot is the reference image corresponding to the chosen label. Note that for both baselines and LaRVAE, we use only 1\% ($\eta=0.01$) of the labeled data.
  }
  \label{fig:traversal-synthetic}
\end{figure}

\begin{figure}[hp!]
    \centering
      \begin{subfigure}[b]{\textwidth}
      \includegraphics[width=0.32\linewidth]{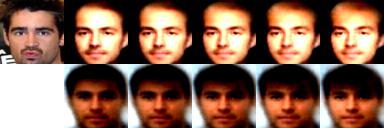}
      \includegraphics[width=0.32\linewidth]{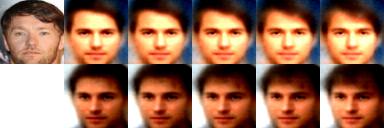}
      \includegraphics[width=0.32\linewidth]{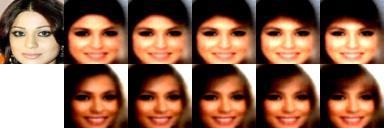}\\
      \includegraphics[width=0.32\linewidth]{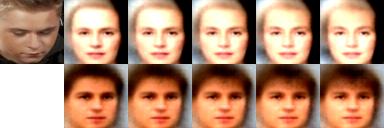}
      \includegraphics[width=0.32\linewidth]{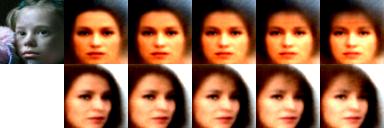}
      \includegraphics[width=0.32\linewidth]{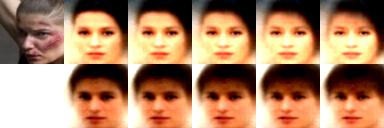}
      \caption{Bangs}
      \end{subfigure}
      \begin{subfigure}[b]{\textwidth}
      \includegraphics[width=0.32\linewidth]{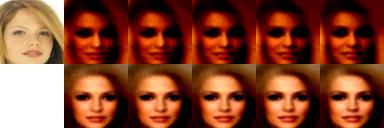}  
      \includegraphics[width=0.32\linewidth]{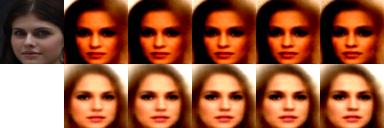}
      \includegraphics[width=0.32\linewidth]{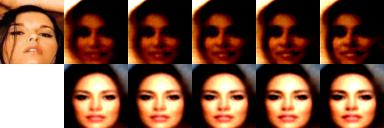}\\
      \includegraphics[width=0.32\linewidth]{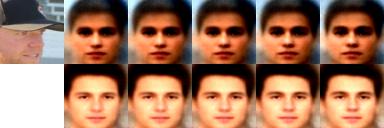}
      \includegraphics[width=0.32\linewidth]{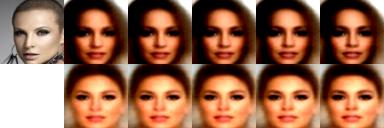}
      \includegraphics[width=0.32\linewidth]{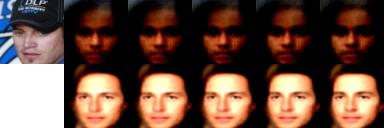}
      \caption{Bushy Eyebrows}
      \end{subfigure}
      \begin{subfigure}[b]{\textwidth}
      \includegraphics[width=0.32\linewidth]{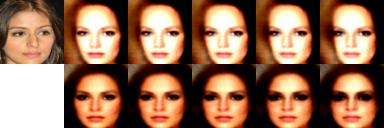}
      \includegraphics[width=0.32\linewidth]{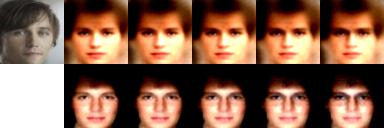}
      \includegraphics[width=0.32\linewidth]{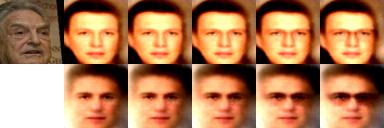}\\
      \includegraphics[width=0.32\linewidth]{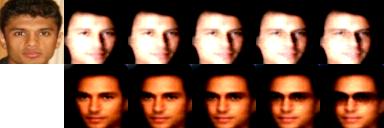}
      \includegraphics[width=0.32\linewidth]{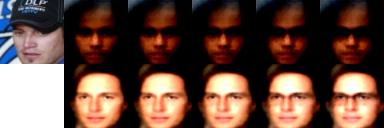}
      \includegraphics[width=0.32\linewidth]{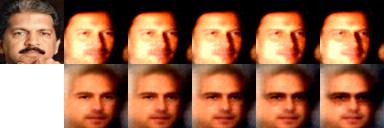}
      \caption{Glasses}
      \end{subfigure}
      \begin{subfigure}[b]{\textwidth}
      \includegraphics[width=0.32\linewidth]{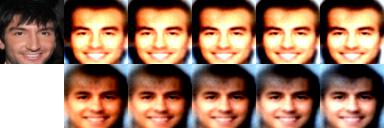}
      \includegraphics[width=0.32\linewidth]{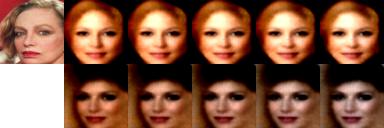}
      \includegraphics[width=0.32\linewidth]{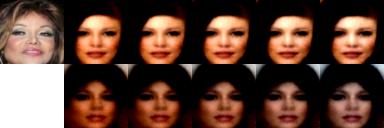}\\
      \includegraphics[width=0.32\linewidth]{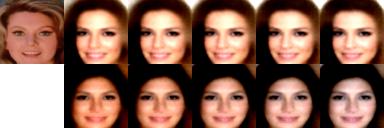}
      \includegraphics[width=0.32\linewidth]{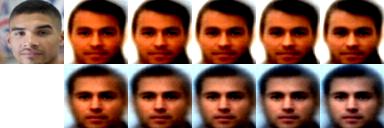}
      \includegraphics[width=0.32\linewidth]{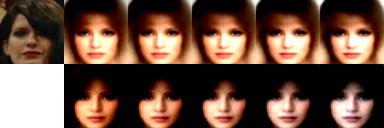}
      \caption{Pale Face}
      \end{subfigure}
      \begin{subfigure}[b]{\textwidth}
      \includegraphics[width=0.32\linewidth]{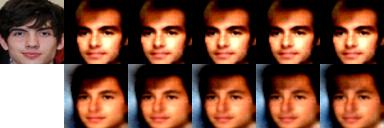}
      \includegraphics[width=0.32\linewidth]{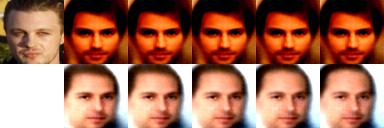}
      \includegraphics[width=0.32\linewidth]{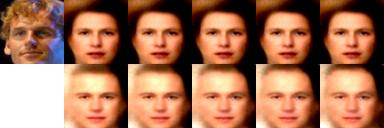}\\
      \includegraphics[width=0.32\linewidth]{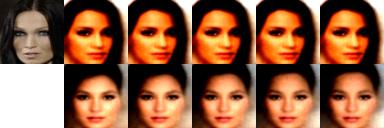}
      \includegraphics[width=0.32\linewidth]{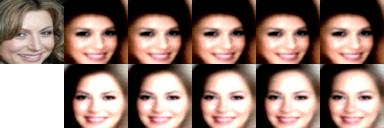}
      \includegraphics[width=0.32\linewidth]{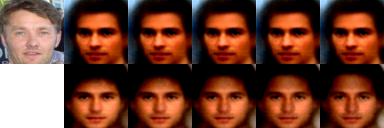}
      \caption{Receding Hairline}
      \end{subfigure}
      \begin{subfigure}[b]{\textwidth}
      \includegraphics[width=0.32\linewidth]{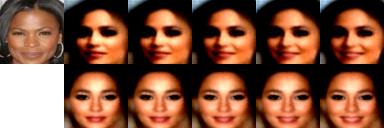}
      \includegraphics[width=0.32\linewidth]{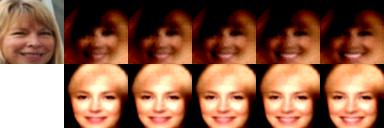}
      \includegraphics[width=0.32\linewidth]{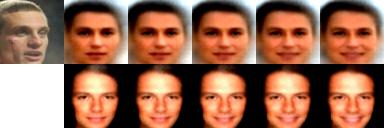}\\
      \includegraphics[width=0.32\linewidth]{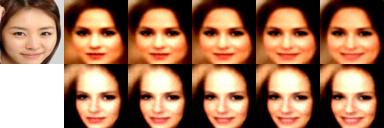}
      \includegraphics[width=0.32\linewidth]{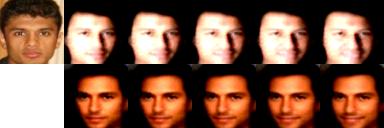}
      \includegraphics[width=0.32\linewidth]{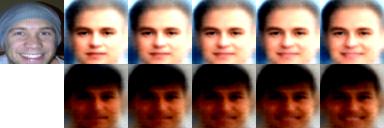}
      \caption{Smiling}
      \end{subfigure}
    \caption{\small Label Traversal visualizations on 6 different label dimensions on the CelebA dataset comparing baseline SS-$\beta$-TCVAE (top row in each plot) and LaRVAE (bottom row in each plot). Leftmost image in each plot is the reference image corresponding to the chosen label. Note that for both baselines and LaRVAE, we use only 1\% ($\eta=0.01$) of the labeled data.}
    \label{fig:traversal-celeba}
\end{figure}
\paragraph{Additional Label Traversal Visualizations}
We present a number of additional label traversal visualizations on the three synthetic datasets in Figure~\ref{fig:traversal-synthetic} and on the real dataset in Figure~\ref{fig:traversal-celeba}. Same as Figure 4 in the main paper, the leftmost image is the reference image corresponding to the label to be traversed. The top row in each plot shows the traversed images generated by the baseline SS-$\beta$-TCVAE and the bottom row shows the traversed images generated by LaRVAE. 

We see that, most of the time, LaRVAE 1) disentangles the specified label dimension better than the baseline and 2) keeps the other factors of variation in the generated images the same as those in the reference image. There are some non-idealities. For example, LaRVAE sometimes fail to maintain the factors of variations specified by the non-traversed dimension in the chosen label. This can be observed, for example, from the bottom left plot in Figure E.3b where the x and y location of the shape is incorrect or from the bottom left plot in Figure E.3c where the wall color (pink instead of green) is incorrect. Nevertheless, overall, LaRVAE generates images that are more visually disentangled than those generated by the baseline while leaves room for improvements.




\end{document}